\documentclass[10pt,twocolumn,letterpaper]{article}

\usepackage{iccv}
\usepackage{graphicx}
\usepackage{amsmath}
\usepackage{amssymb}
\usepackage{booktabs}
\usepackage{enumitem}
\usepackage{tabularx}
\usepackage{subcaption}
\usepackage{soul}
\usepackage{multirow}
\usepackage[table,xcdraw,dvipsnames]{xcolor}
\usepackage{dblfloatfix} 
\usepackage{multirow, adjustbox}
\usepackage{microtype} 
\usepackage[accsupp]{axessibility} 

\usepackage[pagebackref=true,breaklinks=true,letterpaper=true,colorlinks,bookmarks=false]{hyperref}
\usepackage[capitalize]{cleveref}
\crefname{section}{Sec.}{Secs.}
\Crefname{section}{Section}{Sections}
\Crefname{table}{Table}{Tables}
\crefname{table}{Tab.}{Tabs.}

\iccvfinalcopy 

\def\iccvPaperID{9727} 

\ificcvfinal\fi

\usepackage[dvipsnames]{xcolor}
\definecolor{light_orange}{RGB}{255, 229, 204} 
\definecolor{limegreen}{rgb}{0.70, .960, 0.66}
\definecolor{lightgray}{rgb}{0.83, 0.83, 0.83}

\definecolor{csti}{RGB}{108, 142, 191} 
\definecolor{cstii}{RGB}{214, 182, 86} 
\definecolor{cstiii}{RGB}{150, 115, 166} 

\begin{document}

\title{SC3K: Self-supervised and Coherent 3D Keypoints Estimation \\ from Rotated, Noisy, and Decimated Point Cloud Data} 

\author{Mohammad Zohaib, Alessio {Del Bue}\\
Pattern Analysis \& Computer Vision (PAVIS) \\ Italian Institute of Technology (IIT), Genoa, Italy\\
{\tt\small \{mohammad.zohaib, alessio.delbue\}@iit.it}
}

\maketitle
\ificcvfinal\thispagestyle{empty}\fi

\begin{abstract}
    This paper proposes a new method to infer keypoints from arbitrary object categories in practical scenarios where point cloud data (PCD) are noisy, down-sampled and arbitrarily rotated. Our proposed model adheres to the following principles: i) keypoints inference is fully unsupervised (no annotation given), ii) keypoints position error should be low and resilient to PCD perturbations (robustness), iii) keypoints should not change their indexes for the intra-class objects (semantic coherence), iv) keypoints should be close to or proximal to PCD surface (compactness). We achieve these desiderata by proposing a new self-supervised training strategy for keypoints estimation that does not assume any a priori knowledge of the object class, and a model architecture with coupled  auxiliary losses that promotes the desired keypoints properties. 
    We compare the keypoints estimated by the proposed approach with those of the state-of-the-art unsupervised approaches. 
    The experiments show that our approach outperforms by estimating keypoints with improved coverage ($+9.41\%$) 
     while being semantically consistent ($+4.66\%$) that best characterizes the object's 3D shape for downstream tasks. 
    Code and data are available at: \href{https://github.com/IIT-PAVIS/SC3K}{https://github.com/IIT-PAVIS/SC3K} 
\end{abstract}
%
%
%
%
\begin{figure}[t]
\centering
\includegraphics[height=4.8cm]{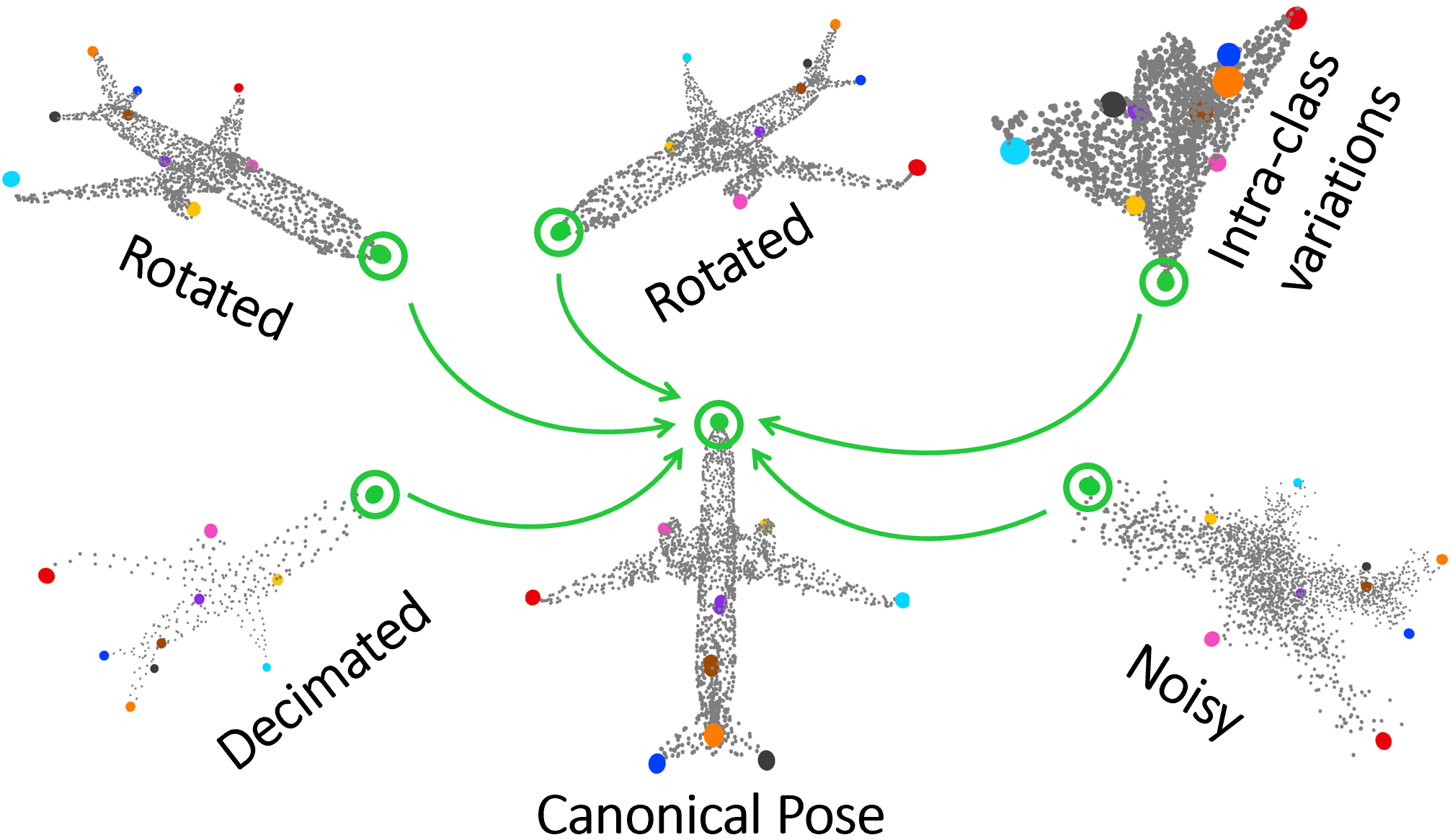}
\caption{Self/un-supervised keypoints estimation from PCD has to be robust to perturbations such as rotations, intra-class shape variations, noisy data and an arbitrary number of input 3D points. 
The keypoint localization needs to be not only accurate and pertain to the object surface, but also preserve semantic coherence, as the green keypoint is always associated with a specific object region despite arbitrary variations in the PCD.
}
\label{fig:over_view}
\end{figure}

\section{Introduction}
\label{sec:intro}
Representing 3D objects using a set of keypoints~\cite{ref:chen2020unsupervised,ref:li2019usip,ref:tang2021multi} is a common and fundamental step for several geometrical reasoning tasks, including pose estimation, action recognition, object tracking, shape registration, deformation, retrieval and reconstruction~\cite{ref:skeleton_marger,ref:you2022ukpgan,ref:wang2018learning,ref:zhong2022snake,ref:jakab2021keypointdeformer}. As being a first processing step, it is crucial that the keypoints (see Fig. \ref{fig:over_view}) are extracted reliably from point cloud data (PCD) of object shapes, as any error may negatively impact further higher-level tasks.

The solution to this problem was initially cast as a supervised learning task: given a dataset of manually annotated PCDs with keypoints, a computational model infers the keypoints position given a PCD as input~\cite{ref:wei2021multi,ref:zoh20223d,ref:liu2020keypose,ref:iqbal2021kama,ref:keypointnet_2020}. While these methods provided impressive results on the dataset they were trained on, they also highlighted the limitations of supervised approaches. The basic issue is the requirement of having large enough datasets containing well-defined ground truth annotations for every object. Annotating such datasets is difficult as finding 3D keypoints manually is a hard and time consuming activity.
Similarly, noise or missing data on the PCD can compromise quality, and highly symmetric/smooth objects might confuse the annotator in finding the correct keypoints. 

Considering such limitations, recent methods have focused on not-supervised approaches to bypass the need for human annotations. Self-supervision methods define proxy tasks for which a large number of annotations can be obtained during training ~\cite{ref:wang2019prnet,ref:poursaeed2020self, ref:you2022ukpgan,ref:chen2022novel,ref:yuan2021self}, e.g., geometrical transformations, canonical mapping, reconstruction to learn the prototype of intra-class object,  etc.~\cite{ref:sajnani2022condor,ref:zhao20223dpointcaps++,ref:sahin2022cmd,ref:sun2021canonical,ref:tang2022prototype}. Unsupervised approaches differently promotes keypoints that are implicitly given by reasoning on the object geometry, e.g. point-level clustering, object's skeleton,  consistency between object's symmetry, part contrasting, etc.~\cite{ref:mei2022unsupervised,ref:xue2021omad,ref:sipiran2011harris,ref:jakab2021keypointdeformer,ref:yuan2022unsupervised}. 

The shift towards these learning paradigms clearly allows generalizing keypoint extraction but not without drawbacks. Without human annotations, it is difficult to identify a specific keypoint in a particular semantic 3D region 
when intra-class variations are present (airplane example in Fig.~\ref{fig:over_view}). 
Moreover, for several applications such as shape registration, it is paramount to maintain the semantic consistency of keypoints, i.e., their vector ordering.  
Despite these considerations, keypoints extraction has to be robust against common perturbations of PCDs, 
and the accuracy in localizing the keypoints should be preserved even if PCDs are rotated, noisy and decimated as shown in Fig.~\ref{fig:over_view}. 

To this end, we propose an approach that reduces the requirement of ground truth labels by utilizing the input PCDs to learn to produce 3D keypoints on the object's surface. It generates two versions of an object by applying a random rotation (as done on images in the methods presented in~\cite{ref:hung2019scops,ref:lorenz2019unsupervised})
and estimates the corresponding keypoints set.

Initially, the network optimizes the keypoints of the individual objects to promote non-overlapping, proximal to the input PCD, and covering the complete object. 
To ensure consistency in the semantic coherence (order) and positions of the estimated keypoints, the network compares the keypoints of both versions of the input PCD.
First, both sets are transformed to the canonical pose and are compared one-to-one between the corresponding keypoints of the sets. 
Second, as a proxy task, the relative pose between the two sets of keypoints is estimated and minimized against the known relative pose of the PCDs pair. 
Such learning strategy and network architecture promote the inference of keypoints that are semantically coherent, robust to perturbations, and with better accuracy.  
 
The main contributions of this work are as follows:
\begin{itemize}[noitemsep,topsep=0pt]
    \item The proposed approach estimates 3D keypoints (from a single PCD), without the need to pre-align a PCD to a canonical pose;
    \item The presented mutual 
    learning procedure allows to estimate keypoints that are semantically consistent for intra-class objects regardless of perturbations, such as rotation, noise, or down-sampling;
    \item  On an average, the presented approach outperforms the state-of-the-art (SOTA) approaches (coverage: +9.41\%, semantic consistency: +4.66\%) and is able to generalize to novel object poses.
\end{itemize}
The rest of the paper is organized as follows; Section~\ref{sec:literature} presents recent keypoints estimation approaches along with their positive features and limitations, Section~\ref{sec:proposed_approach} describes the proposed approach which is evaluated in Section~\ref{sec:results}, Section~\ref{sec:Ablations} reports the ablations, finally conclusions are given in Section~\ref{sec:Conclusions}.

\begin{figure*}[ht]
\centering
\includegraphics[height=7.9cm]{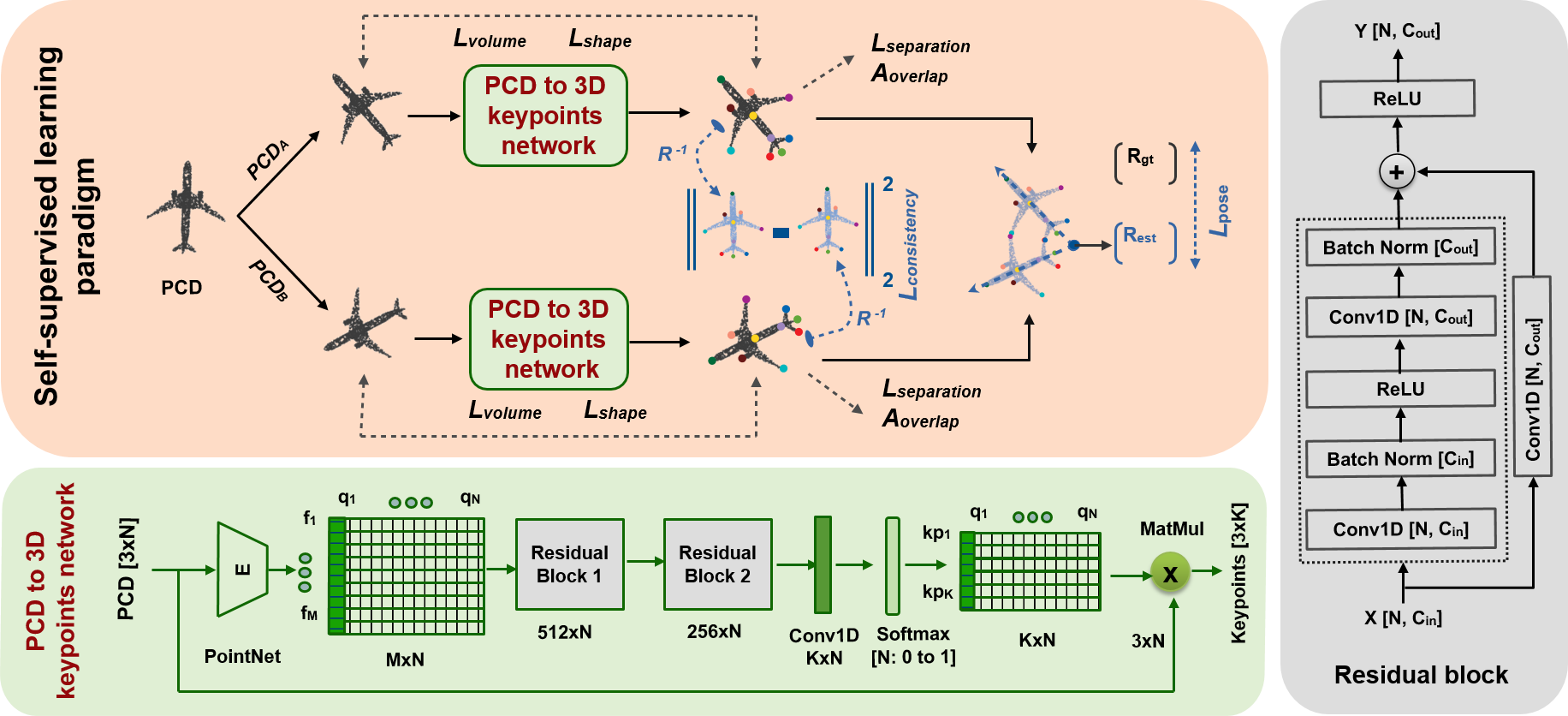}
\caption{PCD to 3D keypoints network (light lime block) takes a PCD of $N$ points as input and extracts \textit{M} global features using PointNet encoder. The features are passed by two cascaded residual blocks (gray block) followed by a convolutional and a softmax layer in order to estimate $K \times N$ probabilities that are used to estimate $K$ keypoints.
The light orange block shows the proposed self-supervised learning paradigm. The method first estimates keypoints for two randomly rotated versions of the same PCD, and then uses them to minimize the individual (volume, shape, separation and average overlap) and mutual (consistency and pose -- as highlighted in \textcolor{NavyBlue}{navy blue}) losses. 
}
\label{fig:complete_architecture}
\end{figure*}
\section{Related works}
\label{sec:literature}
Several methods have been proposed to estimate 3D keypoints in a supervised way using human-annotated keypoints~\cite{ref:wei2021multi,ref:he2020pvn3d,ref:zhang2022pd,ref:lin2022task,ref:zoh20223d,ref:lin2022e2ek,ref:keypointnet_2020}. As our approach is unsupervised, here we review in more detail the methods that do not use supervision. 

Chen et al.~\cite{ref:chen2020unsupervised} present an approach that learns to identify semantically consistent points in the same category in an unsupervised way from an object's PCD. Their network is based on PointNet++~\cite{ref:qi2017pointnet++} that assigns a probability (of being a keypoint) to each element of the PCD. The final keypoints are computed using a convex combination of the points weighted by the probabilities.
Yuan et al.~\cite{ref:yuan2022unsupervised} present an approach that uses two different objects of the same category to estimate semantically ordered 3D keypoints.
Another similar approach that finds correspondences between different objects of the same category is presented in~\cite{ref:cheng2021learning}. 
Li et al.~\cite{ref:li2019usip} present a similar approach that first generates another variant of the PCD by random transformation and then utilizes both PCDs for estimating the keypoints. Their network first generates clusters from the input point clouds and then it estimates a keypoint for every cluster.
A similar approach is presented by Sun et al.~\cite{ref:sun2021canonical}. 
Their network takes two randomly rotated versions of a PCD and computes \textit{K} capsules containing the attention mask for every point in the input PCD and the corresponding features. Based on the attention masks, points are arranged to \textit{K} parts of the object.
Fernandez et al.~\cite{ref:clara_2020} present an approach that estimates symmetric 3D keypoints from PCD. The network estimates \textit{N} nodes 
and applies nonmax-suppression for selecting the final keypoints. However, the approach is very sensitive to object symmetry; thus, its performance may decrease for irregular shapes, i.e., airplanes or guitars, whose geometries vary consistently within the category~\cite{ref:skeleton_marger}. 
The authors in~\cite{ref:skeleton_marger} present ``Skeleton Merger'' (SM) to detect aligned and semantic keypoints from PCDs in an unsupervised fashion. It uses the keypoints to generate a skeleton of the object. Both keypoints and the skeleton are used to reconstruct the PCD. 
Xue et al. present USEEK~\cite{ref:xue2022useek}, a teacher-student network that estimates an equivariant set of 3D keypoints from point clouds. Their teacher module is the same as~\cite{ref:skeleton_marger} which is based on the PointNet++, whereas the student module uses a SPRIN/Vector Neuron SE(3)-invariant backbone. Their network first generates pseudo labels that are required later to train the student module.
Tang et al. propose ``LAKe-Net''~\cite{ref:tang2022lake} that uses the keypoints for the shape completion task. It localizes the aligned keypoints, generates surface-skeleton using the keypoints, and uses them to refine the object's shape. 
Suwajanakorn et al. present an approach~\cite{ref:suwajanakorn2018discovery} to estimate 3D keypoints in the form of 2D positions and depth from a pair of images. Their approach forces 2D keypoints to be
estimated within the object silhouette and uses known camera projections. 
They use their keypoints to estimate the relative pose between objects. 

Considering the limitations of the above-reported literature, this work presents an end-to-end architecture that does not require ground truth labels/silhouettes; rather, it utilizes mutual consistency (relative pose/order) between two versions of the same object as a proxy task to improve the 3D position and semantic coherence of the estimated keypoints. Furthermore, our design and selection of the loss functions allow the keypoints to be estimated proximal to the object's surface.

\section{Proposed approach - SC3K}
\label{sec:proposed_approach}
Given a PCD of an object, the goal of the proposed approach, named SC3K, is to estimate keypoints that are semantically coherent and accurate despite arbitrarily rotated PCDs and perturbations without requiring ground truth annotations. The architecture of the SC3K is illustrated in Fig.~\ref{fig:complete_architecture}.
In the following subsections, we describe the network to estimate 3D keypoints from the PCDs, and our self-supervised learning paradigm.

\subsection{PCD to 3D keypoints network 
(\colorbox{limegreen}\;)
}

The PCD to 3D keypoints network (light lime block in Fig.~\ref{fig:complete_architecture}) uses a PointNet~\cite{ref:qi2017pointnet} backbone to extract \textit{M} features for every point in the input PCD. The extracted features pass through  two consecutive residual blocks that reduce the features from \textit{M} to 256. 
Each residual block (gray block (\colorbox{lightgray}\;) in Fig.~\ref{fig:complete_architecture}) contains a pair of Conv1D layers with batch normalization connected via ReLU, and a skip connection with a single Conv1D layer.
The refined features are later projected to a conv1D and a softmax layer to estimate $K \times N$ features, where \textit{K} represents the total number of keypoints and \textit{N} represents the weight/probability for every point in the input point cloud to be selected as the keypoint.
The weights of every keypoint (\textit{$N\times$1}) are multiplied to the input PCD (\textit{$3\times N$}) in order to estimate the final keypoint (\textit{$3\times1$}). The final keypoint represents a weighted average point of the PCD. We repeat this process \textit{K} times to estimate all the keypoints ($3\times K$).
\subsection{Self-supervised learning paradigm (\colorbox{light_orange}\;)}
\label{sec:training}

The proposed learning paradigm (as shown in the light orange block in Fig.~\ref{fig:complete_architecture}) accepts as input an object PCD that is then randomly rotated twice to obtain two PCDs. These PCDs are then processed by the PCD to 3D keypoints network that outputs two sets of keypoints. This pairwise set will be used as a self-supervised signal to enforce keypoints semantic consistency.
For each set of keypoints a loss with four components is computed, based on how well the keypoints fit the shape of the input PCD. We call this loss ``individual loss''. 
Then, the two sets of keypoints from the two randomly rotated PCDs  are used to compute ``mutual dependency loss'' that contains two components.
In the first component, both the keypoints sets are transformed to the (known) canonical pose to compute the one-to-one consistency between the corresponding keypoints.
In the second component, the relative pose of the keypoints are compared with those of the input PCDs to refine the keypoints position and the semantic coherence. 
The network is trained to minimize  both  loss functions. In the following, we present these two losses in detail.
\subsection{Individual loss}

The individual loss 
is computed 
for a single shape $\mathcal{PCD} = [p_1, p_2, ..., p_N] \in \mathbb{R}^{3 \times N}$ and it outputs a set of $K$ keypoints $\mathcal{KP} = \{k_1, k_2, ..., k_K\}, \in \mathbb{R}^{3 \times K}$ with $K \ll N$. The desired properties of the keypoints are that they should be relatively separated, covering as much as possible the whole object's volume while still being close to the PCD, and not overlapping with each other. These properties are described next. 

\textbf{Separation loss}: 
This loss ($\mathcal{L}_{sep}$) 
maximizes the distance of every keypoint ($k_i$) from its 
neighbouring keypoint ($\textrm{\textit{kNN}}(k_i, \mathcal{KP})$) in  $\mathcal{KP}$ thus promoting more spread out configurations of points.
It is defined as: 
\begin{equation}
  \mathcal{L}_{sep} = \frac{1}{ max \left(\frac{1}{K} \sum\limits_{i=1}^{K} \big\| k_i - \textrm{\textit{kNN}}(k_i, \mathcal{KP})\big\|_2 , 0.01 \right)},
 \label{eq:separation_loss} 
\end{equation}
where the term 0.01 is used to avoid the infinite loss value, which can occur if all the keypoints are estimated at the same position.  

\textbf{Shape loss}:
Since $\mathcal{L}_{sep}$ moves away keypoints from their neighbours without any maximum distance limit,  
keypoints might move easily far from the object 
and even further. Therefore, we use the shape loss ($\mathcal{L}_{shape}$) 
that enforces keypoints being closer to the object's shape.
The loss minimizes the distance of every keypoint $k_i$ in $\mathcal{KP}$ from its nearest neighbour point in the input $\mathcal{PCD}$. The loss can be defined as:
\begin{equation}
  \mathcal{L}_{shape} = \frac{1}{K} \sum\limits_{i=1}^{K} \big\| k_i - \textrm{\textit{kNN}}(k_i, \mathcal{PCD})\big\|_2.
 \label{eq:shape_loss} 
\end{equation}

\textbf{Volume loss}:
The $\mathcal{L}_{sep}$ and $\mathcal{L}_{shape}$ losses 
do not consider how the keypoints are distributed over the whole shape of the object.
Therefore to estimate keypoints that cover the entire object, we propose the volume loss as $\mathcal{L}_{volume}$.
The loss computes the difference between the 3D volume of the estimated keypoints with that of the input PCD as:
\begin{equation}
  \mathcal{L}_{volume} = \big\| vol(\mathcal{KP}) - vol( \mathcal{PCD})\big\|,
 \label{eq:volume_loss} 
\end{equation}
where $vol()$ is a function that computes a volume from a set of points in terms of a 3D bounding box enclosing the points~\cite{ref:clara_2020}. To find the difference in volume, we use smooth L1 loss 
as this loss is less sensitive to outliers compared to the MSE loss~\cite{ref:fastrcnn_2015}. 

\textbf{Average overlap}:
To avoid multiple keypoints being estimated at the same 3D position, we compute the average overlap $\mathcal{A}_{overlap}$ among the keypoints as:
\begin{equation}
  \begin{aligned}
  \mathcal{A}_{overlap} = \frac{1}{K^2}\sum\limits_{i=1}^{K} \sum\limits_{j=1}^{K} \left[ \big\| k_i - k_j  \big\|_2 < \tau_1 \right], \quad i \ne j
  \\
    \left[ \big\| k_i - k_j  \big\|_2 < \tau_1 \right] =
    \begin{cases}
      1 & \text{if true} \\
      0 & \text{otherwise}
    \end{cases}
  \end{aligned}
 \label{eq:overlap_loss} 
\end{equation}
where $[.]$ is the Iverson bracket. 
Two keypoints
are considered as overlapping if the Euclidean distance between them is less than the threshold $\tau_1$, which is 0.05. We add this number to the overall individual loss.

The total individual loss can be summarized as a weighted sum of the above loss components;
\begin{equation}
  \begin{aligned}
\mathcal{L}_{individual} = w_{sep} \cdot \mathcal{L}_{sep}+ w_{sh} \cdot \mathcal{L}_{shape} + \\ w_{vol} \cdot \mathcal{L}_{volume} + w_{ovr} \cdot \mathcal{A}_{overlap},
  \end{aligned}
\label{eq:position} 
\end{equation}
where, \{$w_{sep}$, $w_{sh}$, $w_{vol}$, $w_{ovr}$\} are not optimised hyperparameters fixed to \{0.5,6,1,0.07\} respectively. 

\subsection{Mutual dependency loss}
In order to refine the positions of the keypoints and to make them semantically coherent across different rotations, 
we use the mutual dependency loss. Differently from the individual loss, here we consider the pair of keypoints obtained from the randomly rotated shapes.
The loss is given by two components as described below. 

Suppose that the two randomly rotated versions of the input PCD are {$\mathcal{PCD_A} = [a_1, a_2, ..., a_N] \in \mathbb{R}^{3 \times N} $} and {$\mathcal{PCD_B} = [b_1, b_2, ..., b_N] \in \mathbb{R}^{3 \times N} $} while the \textit{K} keypoints estimated by the proposed approach for each PCD version can be represented as $\mathcal{KP_A}=[k^a_1, k^a_2, ..., k^a_K] \in \mathbb{R}^{3 \times K}$ and $\mathcal{KP_B}=[k^b_1, k^b_2, ..., k^b_K]\in\mathbb{R}^{3 \times K}$, respectively. Then the loss functions can be described as given below.

\textbf{Keypoints consistency loss}:
Consider that $R_a \in \mathbb{R}^{3 \times 3}$ and $R_b \in \mathbb{R}^{3 \times 3}$ are the rotations associated to ${\mathcal{PCD_A}}$ and ${\mathcal{PCD_B}}$, respectively. We use these rotation matrices and transform the keypoints (${\mathcal{KP_A}}$ and ${\mathcal{KP_B}}$) back to their canonical pose. 
The keypoints are said to be coherent if they overlap in this common reference system and if their indexes exactly match. To introduce these desiderata,
we compute the consistency loss ($\mathcal{L}_{consist}$) between the corresponding keypoints in both the transformed sets as:
\begin{equation}
  \begin{aligned}
  \mathcal{L}_{consist} &= \frac{1}{K} \sum\limits_{i=1}^{K} \big\|  R_a^{-1}  {k^a_i} - R_b^{-1} {k^b_i} \big\|^2_2.
  \end{aligned}
 \label{eq:consistency_loss} 
\end{equation}
In this way, we penalize keypoints with the wrong ordering and 3D position errors.

\textbf{Pose loss}:
Our approach also learns to solve an auxiliary and self-supervised keypoints registration task by estimating the rotation matrix that aligns the two sets of keypoints against the (known) rotations in the input PCDs.
Suppose $R_{est}$ is the relative pose between 
$\mathcal{KP_A}$ and $\mathcal{KP_B}$, computed by using orthogonal Procrustes Analysis. Then the pose loss ($\mathcal{L}_{pose}$) can be computed using the Frobenius norm between the $R_{est}$ and relative pose of the PCDs ($R_{ba}= R_a \cdot R_b^T$) as:
\begin{equation}
  \begin{aligned}
  \mathcal{L}_{pose} &= 2 \; arcsin \left( \frac{1}{2\sqrt{2}}  \; || R_{est} - R_{ba} ||_F \right).
  \end{aligned}
 \label{eq:pose_loss} 
\end{equation}
It can be observed that if the keypoints in the canonical pose are not aligned/overlapped, the $R_{est}$  will be erroneous, and hence the loss will be high. In other words, the lower pose loss validates the accuracy of the correspondences in both sets of keypoints.

The mutual dependency loss can be defined as the weighted sum of the above two losses:
\begin{equation}
  \mathcal{L}_{mutual\_dependency} = w_{con} \cdot \mathcal{L}_{consist} + w_{pose} \cdot \mathcal{L}_{pos},
\label{eq:mutual_dependency} 
\end{equation}
where \{$w_{con}$, $w_{pose}$\} are defined as \{1, 0.05\}.
The overall training loss is the sum of the position and the mutual dependency loss;
\begin{equation}
  \begin{aligned}
\mathcal{L}_{overall} = \mathcal{L}_{individual} + \mathcal{L}_{mutual\_dependency}.
  \end{aligned}
\label{eq:overall} 
\end{equation}

\subsection{Implementation details}
\label{sec:implementation_details}
During inference, the proposed approach takes a single PCD as input and estimates a semantically ordered list of \textit{K} keypoints. The rotation of the input PCD can be arbitrary and we do not need any pre-processing step.
The network is implemented in PyTorch and trained using the Adam optimizer with the learning rate $1e^{-3}$. We do not freeze any part of the network. 
In all the experiments, the batch size is set to 32 and trained on a 12GB GPU.
We train~\cite{ref:clara_2020},~\cite{ref:skeleton_marger} and our network for 200 epochs and evaluate them using the best-trained model (with the minimum validation loss).

\section{Experiments and evaluation}
\label{sec:results}
This section presents the dataset, the evaluation metrics, a comparison between our method SC3K and the SOTA approaches, and ablation studies.

\subsection{Dataset}
\label{sec:datasets}
We use KeypointNet dataset~\cite{ref:keypointnet_2020} 
in our experiments, considering that this is the standard and most recent dataset used for keypoints estimation. It contains 8329 objects 
and 83231 keypoints of 16 object categories. We do not use the ground truth keypoints. Whereas, we rotate every object in 24 random poses since during training we need to feed two rotated versions of the same object to the proposed SC3K. 
We use the same rotation matrices that are used in ONet~\cite{ref:mescheder2019onet} with a validation and testing split that differs from the training set.
For a fair comparison, we use the original (not-rotated) dataset to evaluate SC3K and the SOTA approaches.

\subsection{Metrics for unsupervised keypoints estimation}
\label{sec:perofmance_metrics}
To compare the performance of the proposed approach, 
we use three different standard metrics.  
The first metric, \textbf{inclusivity metric}~\cite{ref:clara_2020} computes the percentage of the keypoints ($\mathcal{KP}$), which are estimated close to the $\mathcal{PCD}$. The keypoint ($k_i$) whose distance ($d_i$) to the nearest neighbour point in $\mathcal{PCD}$ is below the predefined threshold ($\tau_2$) is considered as a close keypoint. The metric is defined as:
\begin{equation}
  \begin{aligned}
  d_i &= \big\| k_i - \textrm{\textit{kNN}}(k_i, \mathcal{PCD})\big\|_2 \\
  Inclusivity &= 100 \times \frac{1}{K} \sum\limits_{i=1}^{K} \left[ d_i < \tau_2 \right],
   \end{aligned}
   \label{eq:inclusivity} 
\end{equation}
where $[.]$ is the Iverson bracket (as described in Eq.~\ref{eq:overlap_loss}).
Although the inclusivity loss computes how close the $\mathcal{KP}$ are estimated from the input $\mathcal{PCD}$, it does not evaluate the accuracy of the keypoints in covering the whole object. 
Therefore, evaluation is further supported by the second metric, \textbf{coverage metric} \cite{ref:clara_2020}, which compares the intersection over union of the 3D bounding boxes containing the $\mathcal{KP}$ with that of the $\mathcal{PCD}$. 
The metric is defined as:
\begin{equation}
\begin{aligned}
  Cov &= 100 \times \left[1 - \frac{| vol(\mathcal{PCD}) - vol(\mathcal{KP})|}{vol(\mathcal{PCD})} \right]  \\ 
      Coverage &=
    \begin{cases}
      Cov & \text{if }\; vol(\mathcal{KP}) \leq 2 \times vol(\mathcal{PCD}) \\
      0 & \text{otherwise} ,
    \end{cases}
    \end{aligned}
   \label{eq:coverage} 
\end{equation}
where $vol(.)$ is the function that accepts a set of points ($\mathcal{KP}$ or $\mathcal{PCD}$), identifies a maximum and a minimum point from the accepted set, and returns their difference (i.e., the diagonal distance of the object's bounding box).
The coverage will be 100\% if both bounding boxes fully overlap
and it will decrease if the bounding box of the $\mathcal{KP}$ is either smaller or greater than the one of $\mathcal{PCD}$. 
The third metric, \textbf{Dual Alignment Score (DAS)} evaluates the semantic consistency between the keypoints estimated for different objects of the same category. 
By following the same procedure as given in~\cite{ref:skeleton_marger}, we define the ratio of a set of reference keypoints for each category that are semantically aligned w.r.t. the corresponding human annotated keypoints. 

\subsection{Results and analysis}
\label{sec:resutls_analysis}
We compare SC3K 
with the SOTA approaches ULCS~\cite{ref:clara_2020} and SM~\cite{ref:skeleton_marger} 
that estimate the 3D keypoints in an unsupervised way. We trained and tested them 
using KeypointNet~\cite{ref:keypointnet_2020} dataset, keeping the PCDs in the canonical pose because they do not deal with the random rotation. However, considering the nature of SC3K,
we train it for rotated PCDs (i.e., comparatively a more complex problem).
We test SC3K 
under two conditions: \textit{SC3K$\_{rot}$} (PCDs with the random rotation) and \textit{SC3K$\_{can}$} (PCDs in the canonical pose).
The random rotations are used to evaluate the accuracy of SC3K 
irrespective of the object's pose.
However, to be consistent with our competitors (ULCS and SM), we also test our method for the original PCDs in a canonical pose. 
Tab.~\ref{tab:quantitative_comparison} presents a comparison among ULCS, SM and SC3K 
based on the three performance metrics as discussed in Sec.~\ref{sec:perofmance_metrics}. 
\begin{table*}
\centering
\begin{adjustbox}{width=2.07\columnwidth}
\begin{tabular}{l|cccc|cccc|cccc}
\toprule
\multirow{2}{*}{Category} & \multicolumn{4}{c}{\textbf{Inclusivity $\uparrow$}}                             & \multicolumn{4}{c}{\textbf{Coverage $\uparrow$}}  & \multicolumn{4}{c}{\textbf{DAS $\uparrow$}} \\
                          & \textbf{ULCS} & \textbf{SM}    & \multicolumn{1}{l}{\textbf{SC3K$\_{can}$}} & \multicolumn{1}{l}{\textbf{SC3K$\_{rot}$}} & \textbf{ULCS} & \textbf{SM}    & \multicolumn{1}{l}{\textbf{SC3K$\_{can}$}} & \multicolumn{1}{l}{\textbf{SC3K$\_{rot}$}}  & \textbf{ULCS} & \textbf{SM}    & \multicolumn{1}{l}{\textbf{SC3K$\_{can}$}} & \multicolumn{1}{l}{\textbf{SC3K$\_{rot}$}}
                          \\ \midrule
Airplane                  & 71.02         & 72.05          & \textbf{87.20}                         & \underline{74.30}                                  & 88.63         & 92.59          & \textbf{96.34}                         & \underline{94.37}    & 61.40 & 	77.70  & 	\textbf{82.86}   & \underline{81.32}                          \\
Bed                       & 67.00         & 71.89          & \textbf{80.00}                         & \underline{72.29}                                & \underline{94.17}        & 84.28          & \textbf{98.20}                         & 92.85         & --  &  --  & \textbf{64.87}   & \underline{55.97}                       \\
Bottle                    & 75.44         & 72.84          & \underline{77.36}                                 & \textbf{84.01}                         & 80.93         & 91.44          & \textbf{97.95}                         & \underline{94.16}          & --  &  --  & \textbf{62.73}    & \underline{57.22}                     \\
Cap                       & 57.50         & \underline{59.50}         & 56.25                                  & \textbf{67.14}                         & 60.83         & 85.01          & \textbf{94.64}                         & \underline{91.81}    & --  &        53.00  & 	\textbf{59.72}   & \underline{58.10}                     \\
Car                       & 71.32         & 71.95          & \textbf{76.05}                         & \underline{74.45}                                & 83.69         & \textbf{90.69} & 89.84                                  & \underline{90.19}           & --  &        \textbf{79.40}  &   \underline{75.19}  & 73.81               \\
Chair                     & 68.54         & \underline{69.67}         & 56.65                                  & \textbf{72.33}                         & 83.92         & 85.87          & \textbf{95.31}                         & \underline{90.22}   &    64.30	 & 76.80 & 	\textbf{87.04}    & \underline{86.20}                     \\
Guitar                    & 50.14         & \underline{69.29}         & \textbf{96.47}                         & 69.04                                  & 79.83         & 85.65          & \textbf{97.64}                         & \underline{92.17}      &    -- &  63.10  & 	\textbf{65.67}    & \underline{64.02}                   \\
Helmet                    & 64.10         & \underline{72.41}         & 55.00                                  & \textbf{74.68}                         & 79.87         & 82.09          & \textbf{90.50}                         & \underline{90.44}    & --  &  --  & \textbf{58.55}    & \underline{52.32}                             \\
Knife                     & 52.05         & 92.03          & \textbf{98.33}                         & \underline{ 93.15}                                & 76.84         & 77.39          & \textbf{98.77}                         & \underline{88.77}      & --  &  --  & \textbf{62.98}  & \underline{59.69}                           \\
Motorbike                 & 78.43         & \textbf{95.28} & 85.00                                  & \underline{87.74}                                & 78.87         & 86.12          & \textbf{94.34}                         & \underline{91.33}   & --  &  --  & \textbf{59.41}  & \underline{54.63}                             \\
Mug                       & 47.42         & \underline{65.87}         & 46.25                                  & \textbf{82.37}                         & 89.63         & 83.15          & \textbf{95.15}                         & \underline{91.22}     & --  &    67.20  & 	\textbf{75.25}   & \underline{72.14}                          \\
Table                     & 60.06         & \underline{79.13}         & \textbf{79.15}                         & 73.05                                  & 82.97         & 91.31          & \textbf{97.40}                         & \underline{92.32}            & --  &   70.00  & 	\textbf{76.03}     & \underline{71.62}              \\
Vessel                    & 76.89         & \underline{94.24}          & 92.90                                  & \textbf{95.24}                         & 78.79         & 85.28          & \textbf{97.18}                         & \underline{90.03}                     & --  &  --  & \textbf{75.95}  & \underline{72.19}              \\
Average                   & 64.61         & 75.86          & \underline{75.89}                                 & \textbf{78.44}                         & 81.46         & 86.22          & \textbf{95.63}                         & \underline{91.53}   &   62.85 &  	\underline{69.60} &  	\textbf{69.71}  & 66.09
\\ \bottomrule
\end{tabular}
\end{adjustbox}
\caption{Comparison with the SOTA approaches (ULCS~\cite{ref:clara_2020} and SM~\cite{ref:skeleton_marger}) based on KeypointNet dataset. We test our approach for PCDs in canonical pose (\textit{SC3K$\_{can}$}) and the PCDs rotated in random poses (\textit{SC3K$\_{rot}$}).
The results are calculated for 10 keypoints and the $\tau_2$ for the inclusivity is selected as 0.1. The DAS of ULCS and SM are the same as reported in~\cite{ref:yuan2022unsupervised}, thus we consider only the category available in~\cite{ref:yuan2022unsupervised}. For all the metrics, higher values are best. Bold and underlined numbers represent the first and second best performance, respectively.
}
\label{tab:quantitative_comparison}
\end{table*}
Higher values correspond to better performance for every metric.
The first inclusivity metric shows that, on average, the proposed approach (\textit{SC3K\_rot}) outperforms the SOTA approaches by estimating the keypoints close to the object's surface.
However, \textit{SC3K\_{can}} achieves results better than those of ULCS and comparable to those of SM.  
The metric depends on the total number of keypoints and the tolerance threshold $\tau_2$. We show in the \textcolor{blue}{supplementary material} that inclusivity is high for fewer keypoints and it increases with the increase in the $\tau_2$. 
We select $\tau_2$ as 0.05 and consider 10 keypoints for all the experiments and comparison. 
The second coverage metric shows that on average the proposed approach is successful in estimating the keypoints whose 3D bounding boxes best overlap those of the input PCDs. For all the categories, \textit{Ours\_{can}} achieves better results. 
The third DAS metric validates that on average SC3K 
estimates semantically consistent keypoints. The DAS for the ULCS and SM are the same as reported in~\cite{ref:yuan2022unsupervised}. A detailed table presenting the comparison with the MR~\cite{ref:yuan2022unsupervised} and  ISS~\cite{ref:zhong2009intrinsic} 
is given in the \textcolor{blue}{supplementary material}.
\setlength{\tabcolsep}{4pt}
\begin{table*}[t]

\begin{center}
\begin{adjustbox}{width=\textwidth}
\scriptsize
\begin{tabular}{ccccccccccccccc} 
\toprule
ME &  Airplane & Bed & Bottle & Cap  & Car & Chair & Guitar & Helmet & Knife & Bike & Mug & Table & Vessel & Mean \\
\midrule
$\mu$ & 0.041 & 0.072 & 0.058 & 0.057 & 0.061 & 0.045 & 0.047 & 0.071 & 0.055 & 0.072 & 0.039 & 0.072 & 0.040 & 0.056 \\
$\sigma$ & 0.019 & 0.057 & 0.056 & 0.038 & 0.042 & 0.021 & 0.020 & 0.052 & 0.034 & 0.040 & 0.023 & 0.051 & 0.031 & 0.037  \\
\bottomrule
\end{tabular}
\end{adjustbox}
\end{center}
\caption{Pose coherent test: The keypoints estimated for randomly rotated versions of the same object are first transformed to the canonical pose. Then ME ($\mu$ and $\sigma$) is computed between the corresponding keypoints.
}
\label{tab:coherence}
\end{table*}

\begin{figure}[]
\centering
\includegraphics[height=5.0cm]{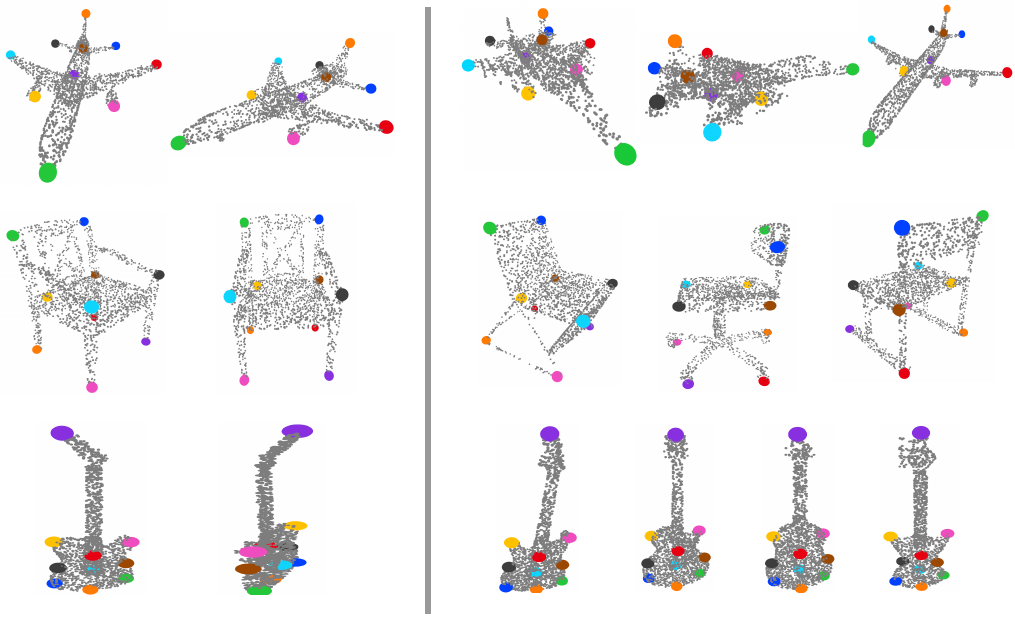}
\caption{Shape pose variations vs. semantic correspondence: columns 1,2) keypoints estimated for two rotated versions of the same object are pose coherent; columns 3-5) keypoints semantically correspond to intra-class variations, in the same they also correspond to those estimated for different objects of the same category.
}
\label{fig:semantic_comparison}
\end{figure}

Unlike the existing approaches, we also evaluate the coherence property of the keypoints by computing the Matching Error (ME). 
This error is a localization error of the keypoints given PCD perturbations. We first estimate keypoints for different rotated versions of the same object PCDs and transform them to the canonical pose using the known rotations. Since the estimated keypoints are in the correct order, we compute order-wise position error between the corresponding keypoints on the canonical reference frame. A low error would indicate that 2D projection of a keypoint is rather unaffected by variations of the PCD.
We repeat this procedure for all the instances of a category and calculate the ME in terms of mean error ($\mu$) and the standard deviation ($\sigma$). 
The quantitative results are depicted in Tab.~\ref{tab:coherence}.
The qualitative results of this experiment are illustrated in Fig.~\ref{fig:semantic_comparison}. 
Where, columns 1 and 2 (on the left side) show the keypoints estimated for two transformed versions of the same objects. It can be seen that the corresponding keypoints are semantically consistent irrespective of the object's pose, this validates the keypoints are coherent.
The keypoints on the right side of the same Fig.~\ref{fig:semantic_comparison} (columns 3, 4 and 5) illustrate the keypoints estimated for different rotated objects of the same category. It can be observed that the keypoints also maintain the correspondences across the different intra-class variations of the object class. 
\section{Ablation studies}
\label{sec:Ablations}
This section presents three ablations on: \textit{i)} Choice of individual loss, \textit{ii)} evaluation and performance of the network with combinations of the different training losses; \textit{iii)} effect of varying noise ratio and decimations of the PCDs. Please refer to the \textcolor{blue}{supplementary material} for additional related ablations. 

\subsection{Chamfer Distance (CD) vs. individual losses}\label{sef:ablation_iou_wrt_kp}
Most of the existing approaches (including our competitors~\cite{ref:clara_2020,ref:skeleton_marger}) have used a variant of the CD to train their networks; instead, we use individual losses. 
It is because the individual loss is more controllable, i.e., we can regulate  every component by setting specific weight. This can be validated from the results presented in Tab.~\ref{tab:quantitative_comparison}, i.e., SC3K 
outperforms the~\cite{ref:clara_2020,ref:skeleton_marger}.
Furthermore, we also present an ablation to highlight the significance of our selection. We train our network by replacing the individual losses with the CD loss.
We observed that the keypoints are not estimated on the object's surface, and some are close to each other. 
Also, the inclusivity
and coverage of the model trained with the individual
loss are \textbf{+5.02} and \textbf{+19.83} better than the model trained with CD, respectively.

\begin{figure*}[t]
\centering
\includegraphics[height=2.68cm]{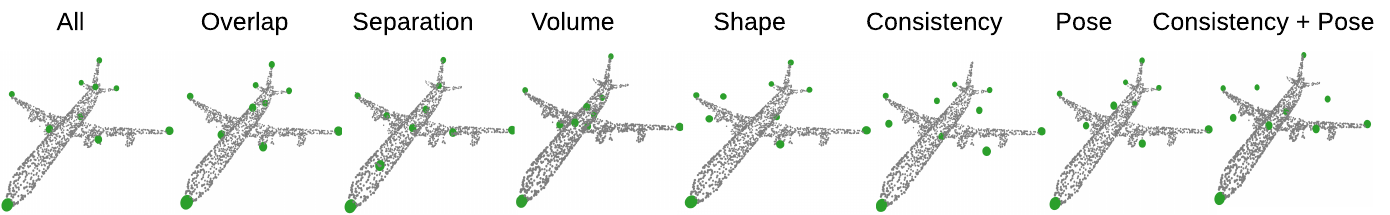}
\caption{Performance of our approach with different combinations of losses. The leftmost figure shows the keypoints when the network is trained for all the losses. In the remaining figures, the model is trained without a specific loss which is mentioned at the top of every figure.}
\label{fig:wo_loss}
\end{figure*}

\subsection{Performance for the selected losses}
\label{sec:selected_loss}
In order to highlight the significance of every loss, 
we train and evaluate the proposed approach by ignoring each loss one by one. The results are illustrated in Tab.~\ref{tab:ablation_loss}. 
The conditional formatting green-to-red shows high-to-low values. 
It can be observed that 
approach performs well overall when all the loss functions are used. The overlap loss contributes comparatively low and is required only at the beginning of the training when the keypoints are estimated randomly. 
The contribution of the separation loss is comparatively higher than the overlap, shape and volume loss since it maintains the distance between the estimated keypoints, thus enforcing the keypoints to move over the whole object and toward the surface. 
Shape loss avoids the estimation of the keypoints outside the object. .
The contribution of the volume loss is comparatively lower than the other loss functions. 
The consistency and pose losses allow the estimation of the corresponding and pose coherent keypoints. Ignoring both the losses 
affects the overall performance of the proposed approach. The qualitative results of the proposed approach trained without the selected loss function are illustrated in  Fig.~\ref{fig:wo_loss}.
\begin{table}[]
\begin{center}
\small
\begin{tabularx}{\linewidth}{Xcccc}
\hline
\noalign{\smallskip}
w.o. loss  & Inc. $\uparrow$ & Cov. $\uparrow$  & DAS $\uparrow$ & ME $\downarrow$ \\
\noalign{\smallskip}
\hline
All loss       & \cellcolor[HTML]{63BE7B}78.44 & \cellcolor[HTML]{63BE7B}91.53 &   \cellcolor[HTML]{63BE7B}74.00 & \cellcolor[HTML]{63BE7B}0.056 \\
$\mathcal{A}_{overlap}$      & \cellcolor[HTML]{D6EDDE}77.09 & \cellcolor[HTML]{FBF9FC}90.72 &  \cellcolor[HTML]{FBE3E6}53.80 & \cellcolor[HTML]{8ECFA0}0.061 \\
$\mathcal{L}_{sep}$     & \cellcolor[HTML]{F8696B}63.01 & \cellcolor[HTML]{F8696B}85.70 & \cellcolor[HTML]{9CD6AD}67.38 & \cellcolor[HTML]{FBD7DA}0.081 \\
$\mathcal{L}_{shape}$   & \cellcolor[HTML]{FBF5F8}76.05 & \cellcolor[HTML]{FBEDF0}90.31 &   \cellcolor[HTML]{E9F5EF}58.45 & \cellcolor[HTML]{A8DAB7}0.064 \\
$\mathcal{L}_{volume}$       & \cellcolor[HTML]{C1E4CC}77.35 & \cellcolor[HTML]{EAF5EF}90.90 & \cellcolor[HTML]{C1E4CC}63.14 & \cellcolor[HTML]{BAE1C6}0.066 \\
$\mathcal{L}_{consist}$      & \cellcolor[HTML]{FBFAFD}76.52 & \cellcolor[HTML]{CFEAD8}91.03 &  \cellcolor[HTML]{F86E70}42.44 & \cellcolor[HTML]{F8696B}0.103 \\
$\mathcal{L}_{pos}$       & \cellcolor[HTML]{F2F8F7}76.76 & \cellcolor[HTML]{CCE9D6}91.04 &   \cellcolor[HTML]{FBE4E7}53.89 & \cellcolor[HTML]{FA9194}0.095 \\
$\mathcal{L}_{consist}$ + $\mathcal{L}_{pos}$  & \cellcolor[HTML]{FAB6B8}70.15 & \cellcolor[HTML]{F9ADAF}88.07 &  \cellcolor[HTML]{F8696B}41.95 & \cellcolor[HTML]{F8696B}0.103 \\
\hline
\end{tabularx}
\end{center}
\caption{Performance of the proposed approach for selected losses where, Inc., Cov., and ME represent inclusivity, coverage and matching error (coherence). 
The conditional formatting ``green-to-red'' represents the ``good-to-bad'' performance.
The results are the average values of the test set of the keypointNet dataset.
}
\label{tab:ablation_loss}
\end{table}
\subsection{Robustness to perturbations}
\label{sec:random_noise_down_sampling}
\begin{figure}
     \centering
      \begin{subfigure}[b]{0.45\textwidth}
         \centering
         \includegraphics[width=7.5cm]{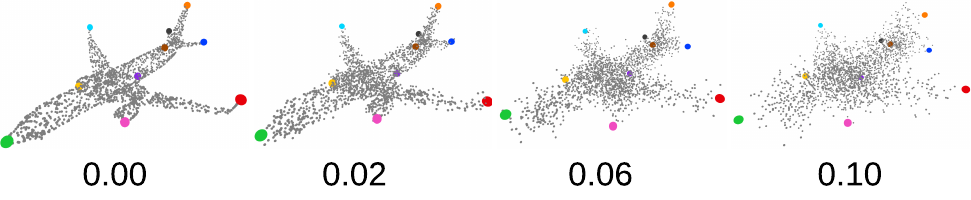}
         \caption{Visualizations of the noisy PCDs}
         \label{fig:noise_comparison_fig}
     \end{subfigure}
     \vfill
      \begin{subfigure}[b]{0.45\textwidth}
         \centering
         \includegraphics[width=7.5cm]{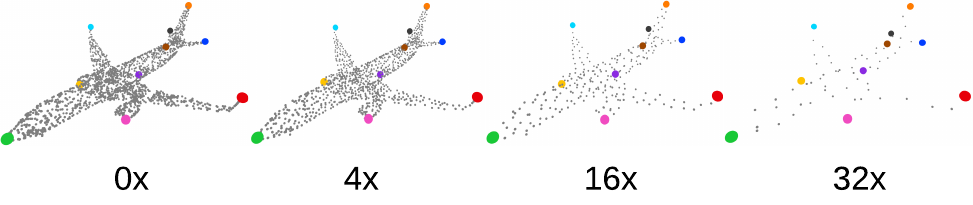}
         \caption{Visualizations of the down-sampled PCDs}
         \label{fig:downsample_comparison_fig}
     \end{subfigure}
     \vfill
     \begin{subfigure}[b]{0.23\textwidth}
         \centering
         \includegraphics[width=4.0cm]{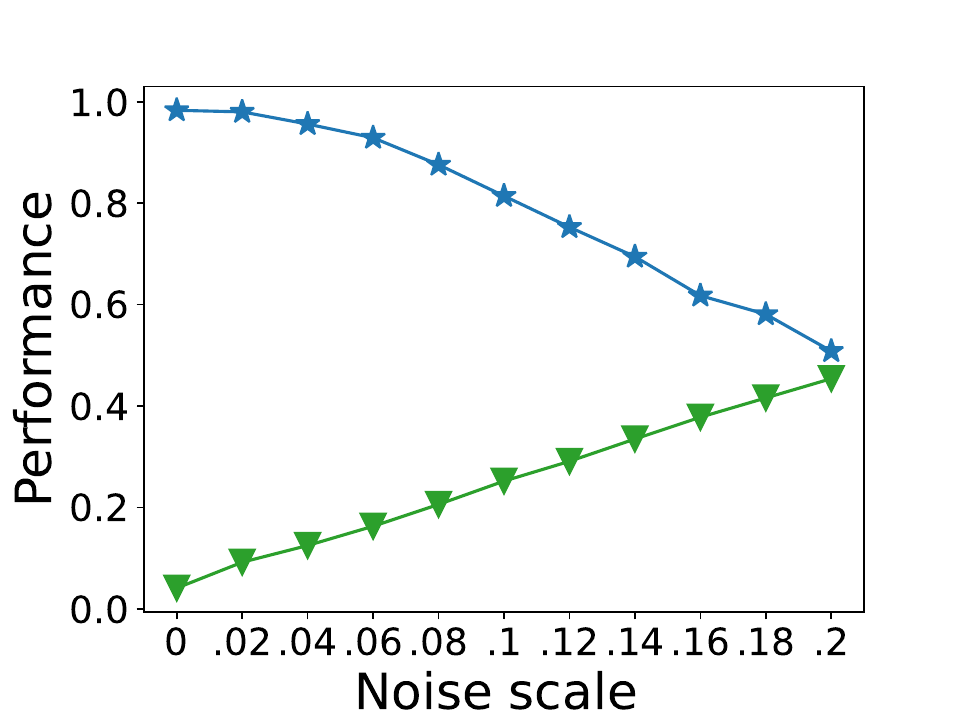}
         \caption{Noise}
         \label{fig:noise_comparison_plot}
     \end{subfigure}
     \hspace{-0.4cm}
     \begin{subfigure}[b]{0.23\textwidth}
         \centering
         \includegraphics[width=4.0cm]{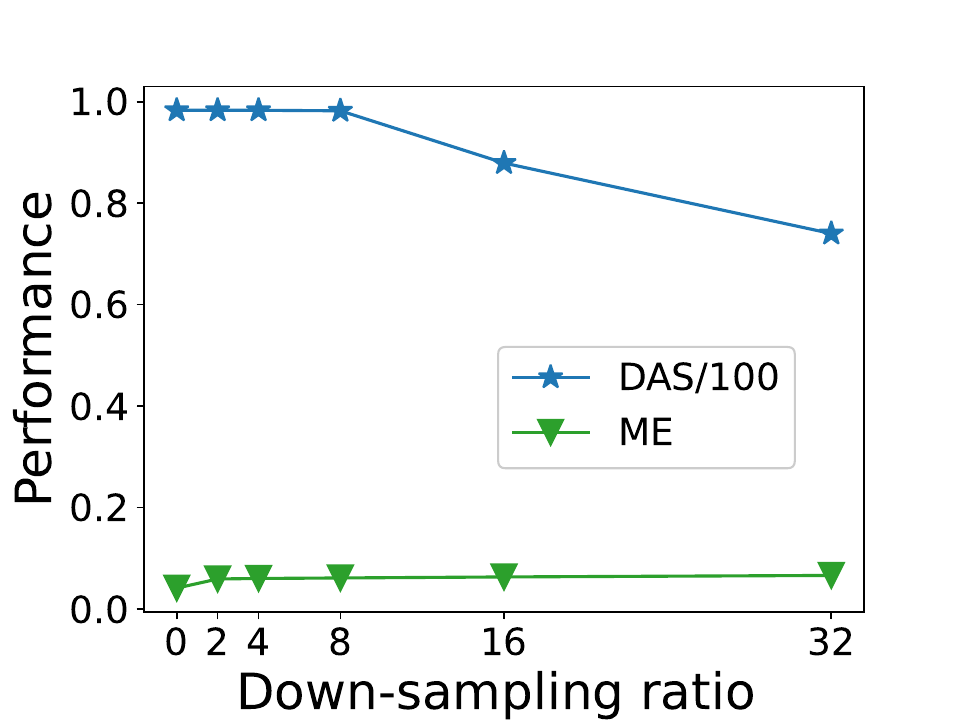}
         \caption{Down-sampling}
         \label{fig:downsample_comparison_plot}
     \end{subfigure}
\caption{Performance of the proposed approach for noisy and decimated PCDs. (a) and (b) represent qualitative results, whereas, (c) and (d) illustrates the effect of the noise scale and down-sampling ratio, respectively.
}
\label{fig:noisy_downsample_comparison}
\end{figure}

This ablation highlights the performance of the proposed approach for noisy and down-sampled PCDs of the airplane category. Noisy PCDs are generated by adding Gaussian noise of different variances to the original PCDs. For decimating the PCD, we use the Farthest Point Sampling (FPS) as used in~\cite{ref:mohammadi2021pointview,ref:yuan2018pcn}.  
Fig.~\ref{fig:noise_comparison_fig} and~\ref{fig:downsample_comparison_fig} show the keypoints estimated for noisy and down-sampled PCDs, respectively. Our approach successfully estimates the consistent keypoints at accurate positions for the noisy and down-sampled PCDs.

Quantitative results for the noisy and down-sampled PCDs are illustrated 
in Fig.~\ref{fig:noise_comparison_plot} and Fig.~\ref{fig:downsample_comparison_plot}, respectively where to fix the DAS in the plots [0 to 1], we show DAS/100. The results show that the ME increases and the DAS decrease with the increase in the noise level. 
Similarly, DAS decreases if down-sampling ratio is reduced to 6 times the original PCD. The ME remains approximately the same for down-sampled PCDs, validating that down-sampling does not affect the keypoints. 
\section{Conclusions}
\label{sec:Conclusions}
This paper presented a method, SC3K, to estimate 3D keypoints from a single PCD such that they express the following properties: {\textit{robust}} -- minimum position error across different rotated versions of the same PCD; {\textit{compact}} --  proximal to the PCD surface, and {\textit{coherent}} -- in semantic order for all the intra-class instances. Similarly, the proposed method is {\textit{repeatable}} -- can estimate the accurate keypoints irrespective of the noise, down-sampling or rotation of the PCD; and {\textit{self-supervised}} -- can estimate the same keypoints from single PCD without requiring any labels (pseudo or human annotation) during the inference.
We achieved these desiderata by training the network with a new self-supervised strategy that does not require human annotations, instead, it computes the relative pose between the two sets of keypoints as a proxy task and then minimizes the error against the known relative pose of the input PCDs pair.
The proposed approach is compared with the SOTA keypoints estimation approaches using the KeypointNet dataset. 
The results validated that the proposed SC3K 
outperforms the SOTA approaches by estimating the coherent keypoints close to the object's surface, characterizing the object's shape.   

\textbf{Limitations.} SC3K may fail to estimate keypoints close to the object's surface for a number of keypoints higher than 35 and its performance decreases for symmetrical shapes. 
For some categories, such as bikes or cars, it is challenging to differentiate between the front and back wheels. 
In the same way, as it happens in previous approaches, strong intra-class geometrical variations negatively affect the performance, i.e., it is hard to compute semantically coherent keypoints between a single and a bunk bed. 
SC3K uses the publicly available dataset and estimates the keypoints to represent an object's shape. So, it does have very limited negative societal impacts. 
\vspace{0.5cm}
\\
\textbf{Acknowledgements:}
We would like to acknowledge Milind Gajanan Padalkar, Matteo Taiana and Pietro Morerio for fruitful discussions, and Seyed Saber Mohammadi and Maryam Saleem for their support during the experimental phase. This work has been supported by the projects ``RAISE-Robotics and AI for Socio-economic Empowerment'' and 
``European Union-NextGenerationEU''. 
%
%
%
{\small
\bibliographystyle{ieee_fullname}
\bibliography{main}
}
%
%
%
%
%
%
%
%
%
%
%
%
%
%
%
%
%
%
%
%
%
%
%
%
%
%
%
%
%
%
%
%
%
%
%
%
%
%
%
%
%
%
%
%
%
%
%
%
%
%
%
%
%
%
%
%
%
%
%
%
%
%
%
%
%
%
%
%
%
%
%
%
%
%
%
%
%
%
%
%
%
%
%
%
%
%
%
%
%
%
%
%
%
%
%
%
%
%
%
%
%
%
%
%
%
%
%
%
%
%
\clearpage
\appendix
\begin{minipage}{0.9\columnwidth}
\centering{\textbf{\LARGE Supplementary Material}}
\vspace{1cm}
\end{minipage}
\section{Introduction}
    This file contains supplementary material for 
    \textit{SC3K: Self-supervised and Coherent 3D Keypoints Estimation
    from Rotated, Noisy, and Decimated Point Cloud Data}. Due to the limited space in the main paper, here we present additional ablations related to the experimental section of the paper, provide a complete table (as given in~\cite{ref:yuan2022unsupervised}) comparing the DAS of our approach with other methods, and show the qualitative results of our experiments.
    Moreover, we also share a video representing 3D visualizations of the results reported in the main paper (i.e., the keypoints estimated by SC3K and SOTA approaches). The video (rotating keypoints) makes it easy to understand the robustness of the SC3K.

\section{Additional Ablations}
This section presents seven additional ablations: 1) comparison of the estimated keypoints with the vertices of a convex hull (CH) of the object, 2) evaluation of SC3K on partial PCDs, 3) performance of SC3K for different number of keypoints, 4) sensitivity of the inclusivity metric with respect to (w.r.t.) the threshold $\tau_2$, 5) impact of the separation and shape loss on the performance of SC3K, 6) impact of the residual block on SC3K, and 7) evaluation of SC3K for different augmentations.

\subsection{Keypoints vs. vertices of a Convex Hull (CH)}
\begin{figure}[h]
  \centering
  \begin{subfigure}{0.45\linewidth}
         \centering
         \includegraphics[width=3.3cm]{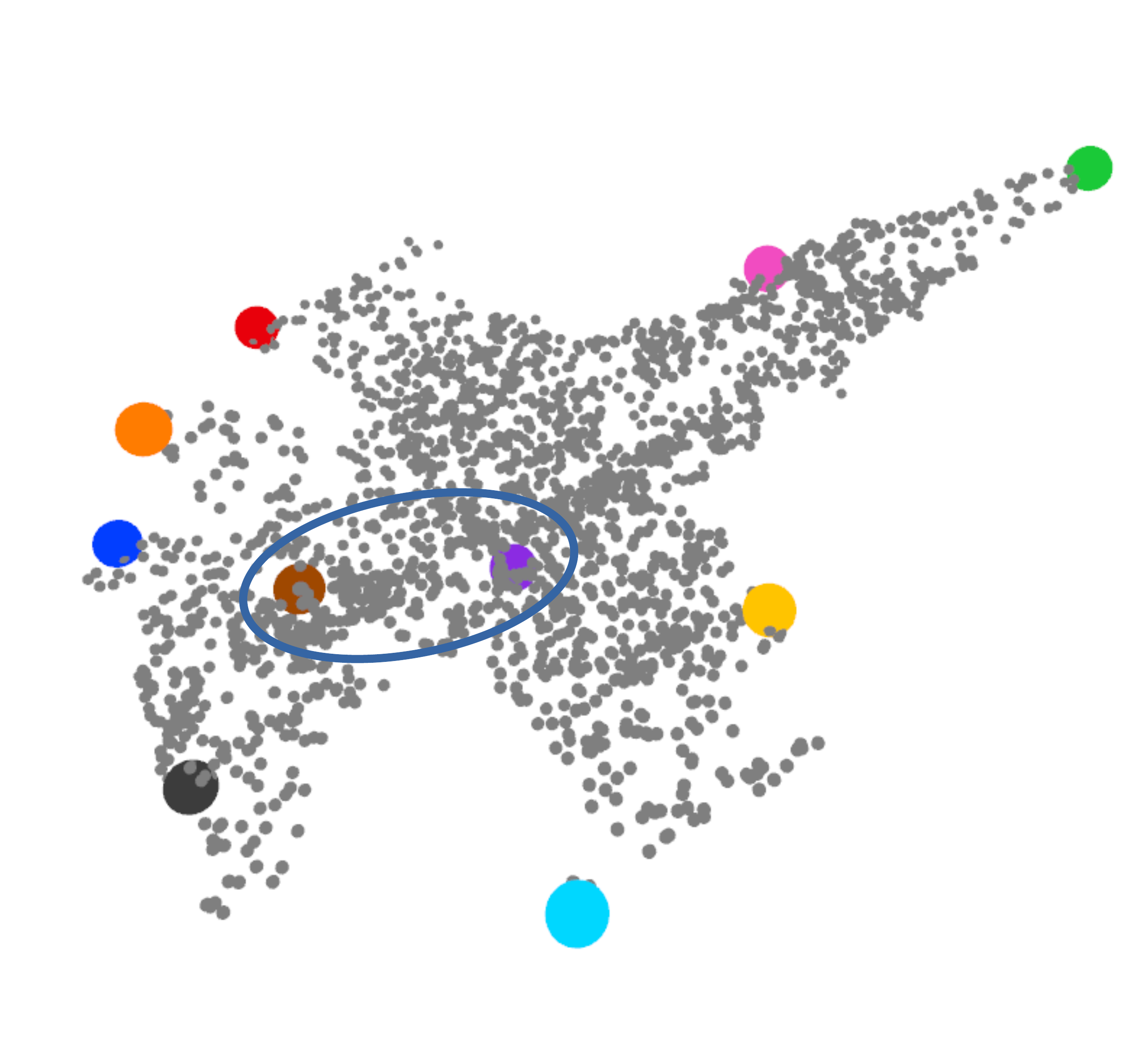}
         \caption{Keypoints of SC3K}
         \label{fig:rebuttal_convexVertices1}
  \end{subfigure}
  \begin{subfigure}{0.45\linewidth}
    \centering
    \includegraphics[width=3cm]{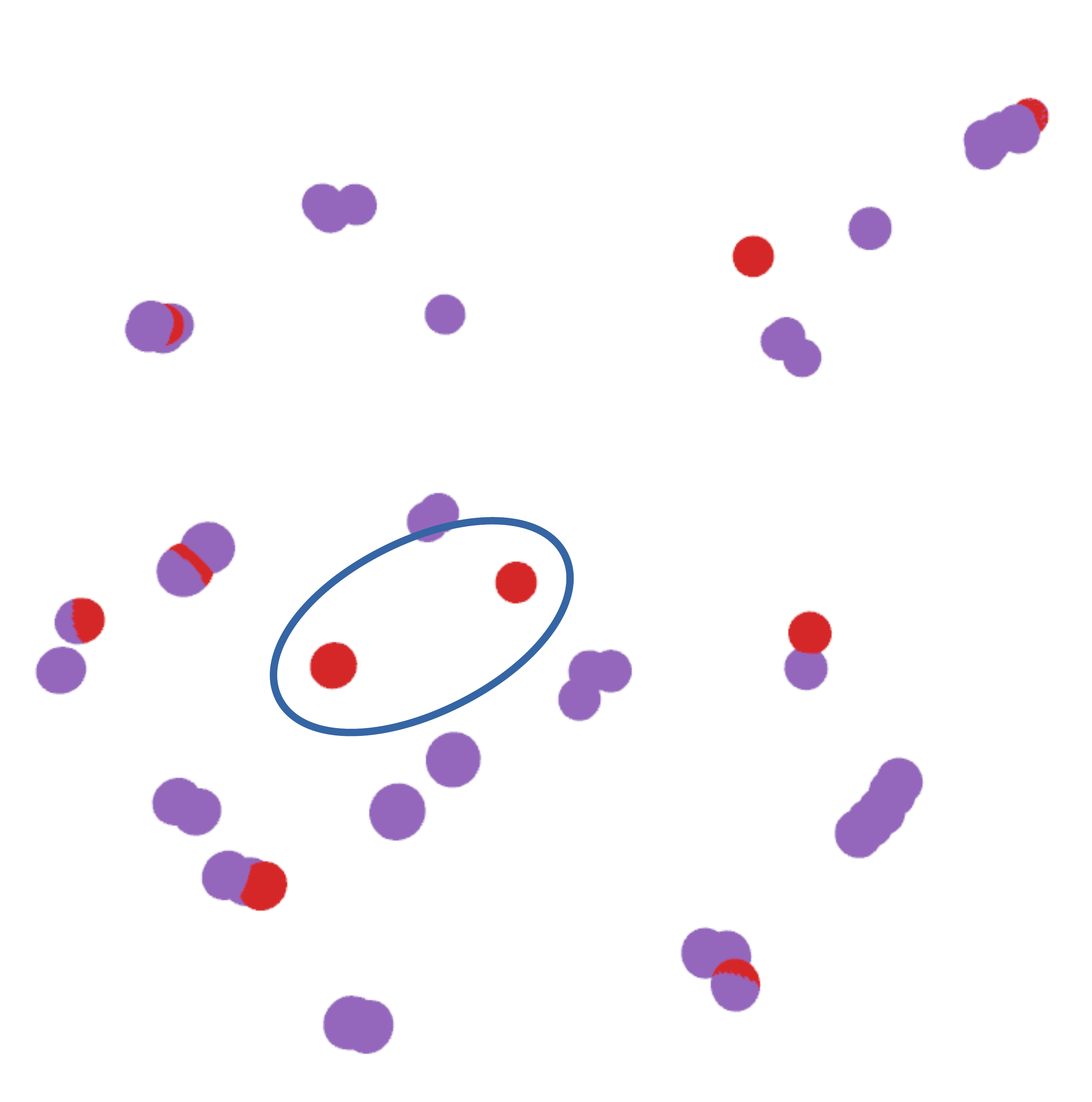}
    \caption{Vertices of the CH}
    \label{fig:rebuttal_convexVertices2}
  \end{subfigure}
  \caption{Comparison: (a) keypoints estimated by SC3K w.r.t. the object, (b) \textcolor{violet}{vertices of the CH} w.r.t. the \textcolor{BrickRed}{keypoints} of the SC3K. Some of the \textcolor{BrickRed}{keypoints} are not included in the \textcolor{violet}{vertices of the CH}.}
  \label{fig:rebuttal_convexVertices}
\end{figure}

Since our approach estimates keypoints on the object's surface, one can relate them with the vertices of the CH of the object. Therefore in this section, we highlight the significance of our keypoints over the vertices of the CH. 
We observed that the vertices of the CH are not as expressive as our semantic keypoints due to the following reasons:
\begin{itemize}
    \item CH vertices do not maintain semantic information (ordering) that represents correspondences between two or more views. To validate this, we implemented a new baseline where we computed vertices of the CH of the test samples, selected 10 meaningful vertices (since 10 keypoints are used for comparison) by using K-means clustering, and computed the DAS metric for them. It is found that the DAS of SC3K (82.86) is \textbf{+68.87} higher than the one obtained from the CH vertices (13.99).
    
    \item CH vertices will not include points that lie within the convex hull (see Fig.~\ref{fig:rebuttal_convexVertices}), which might be important for some downstream tasks, e.g. characterising the shape of objects.
\end{itemize}

\subsection{Evaluation od SC3K on partial PCDs}
\begin{figure}[h]
     \centering
     \begin{subfigure}[b]{0.23\textwidth}
         \centering
         \includegraphics[width=\textwidth]{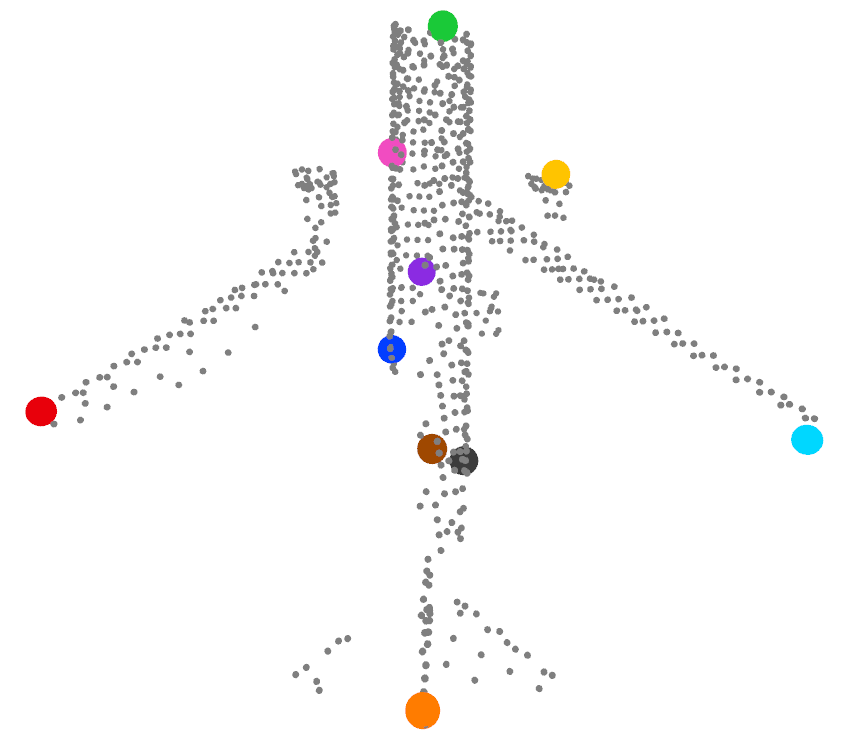}
     \end{subfigure}
     \hfill
     \begin{subfigure}[b]{0.23\textwidth}
         \centering
         \includegraphics[width=\textwidth]{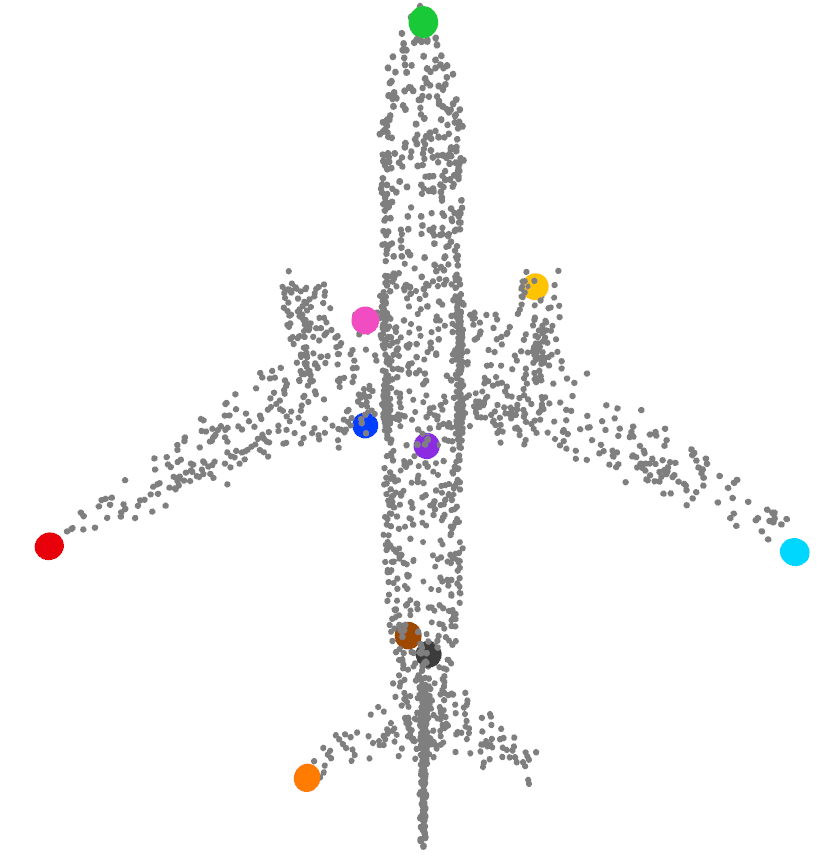}
     \end{subfigure}
     \caption{Visualizations of the keypoints estimated on partial and completed PCD.}
     \label{fig:partial_pcds}
\end{figure}

SC3K is flexible in the backbone used, so it is possible to insert a Point Completion Network (PCN)~\cite{ref:yuan2018pcn} that accepts partial PCDs.
The updated network first estimates the missing parts of the object and then predicts the keypoints.
We use the pretrianed weights of the PCN. We consider common categories of the PCN and keypointNet dataset and select five seen (airplane, car, chair, table, vessel) and three unseen (bed, guitar, motorbike) categories. 
The results for the selected categories are given in Tab.~\ref{tab:partial_pcds}.
As expected, the performance is higher for estimated complete PCDs (called as dense) and lower for partial PCDs. Consider that the updated approach depends on the shape completion module. Thus results are not very impressive where the shapes are not properly estimated. 
\begin{table}[t]
\centering
\begin{adjustbox}{width=0.95\columnwidth}
\begin{tabular}{lcccc}
\toprule 
\multicolumn{1}{c}{\textbf{Data type}} & \textbf{Inclusivity} & \textbf{Coverage} & \multicolumn{1}{l}{\textbf{DAS}} & \multicolumn{1}{l}{\textbf{Average}} \\
\midrule 
\multicolumn{5}{c}{Seen categories}     \\
\midrule 
Partial                                & 46.65                & 89.12             & 54.46                            & 63.41                                \\
Dense                                  & 77.61                & 94.66             & 79.45                            & 83.90                                \\
Complete                               & \textbf{90.69}       & \textbf{95.09}    & \textbf{80.27}                   & \textbf{88.68}                       \\
\midrule 
\multicolumn{5}{c}{Unseen categories}       
\\ \midrule 
Partial                                & 62.78                & 91.40             & 51.92                            & 68.70                                \\
Dense                                  & 48.27                & 94.11             & \textbf{72.73}                   & 71.70                                \\
Complete                               & \textbf{87.11}       & \textbf{96.76}    & 64.27                            & \textbf{82.71}   \\ \bottomrule                 
\end{tabular}
\end{adjustbox}
\caption{Results for the PCN dataset. We test SC3K separately on PCN seen and unseen categories. On average, the results are improved when the input shapes are completed (dense) using the PCN network.}
\label{tab:partial_pcds}
\end{table}
Fig.~\ref{fig:partial_pcds}. 
shows keypoints predicted on partial and estimated complete PCDs.

\subsection{Performance of the proposed approach for different numbers of keypoints}
\begin{figure}[h]
\centering
\includegraphics[height=4.8cm]{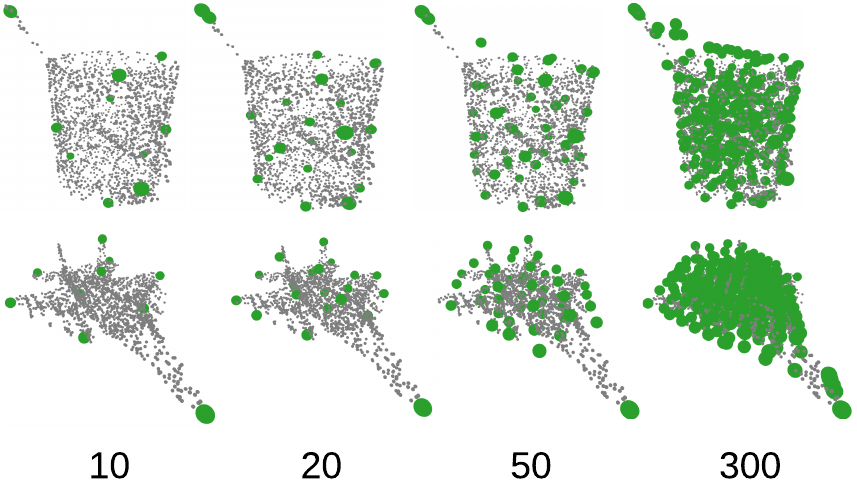}
\caption{Estimation of different numbers of keypoints for the same object. The keypoints are estimated on the object's surface if they are less than or equal to 35 in number. They are predicted outside the object (in case of more than 35 keypoints), especially for a detailed object having empty spaces among its parts.}
\label{fig:diff_kpts}
\end{figure}

We evaluate our approach by varying the number of computed keypoints from the PCD. We found that for most of the shapes (e.g., bottle, guitar), our approach estimates keypoints over the surface of the object. 
However, for the detailed objects with gaps between the parts (i.e., airplanes have relevant empty spaces between a wing and the tail), some of the keypoints are estimated outside the object (in the gaps). 
This effect appears only when a high number of keypoints are considered (higher than $35$).  
As an example, different numbers of keypoints estimated 
for the cup and airplane category
are shown in Fig.~\ref{fig:diff_kpts}.

\subsection{Inclusivity metric and its sensitivity w.r.t. to the parameter $\tau_2$}
\begin{figure}[h]
\centering
\includegraphics[height=6cm]{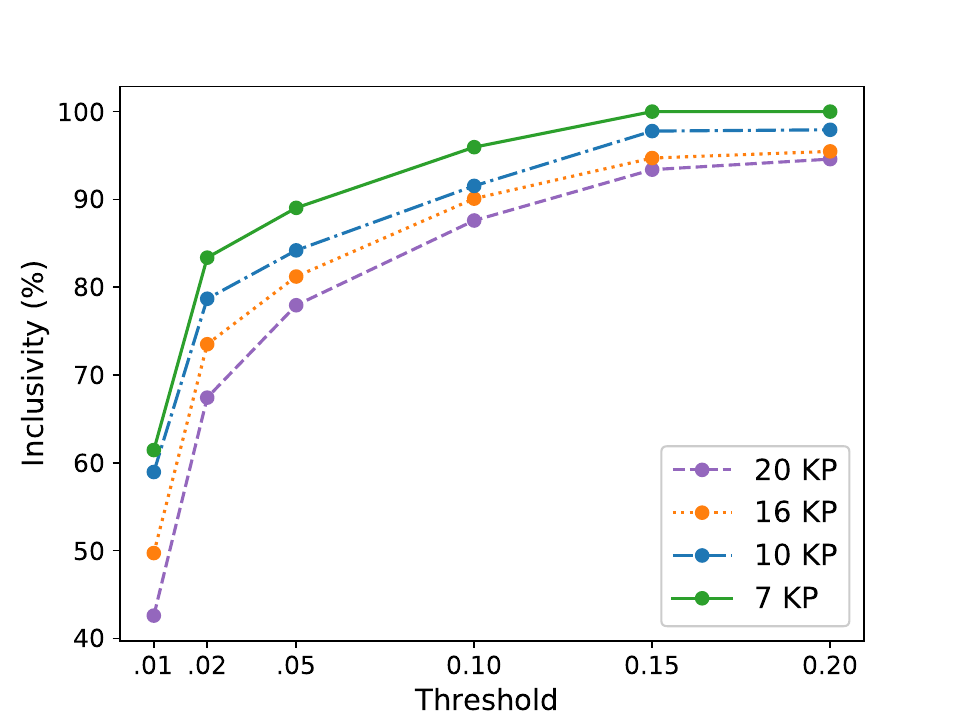}
\caption{Average inclusivity of the proposed approach for different keypoints and thresholds ($\tau_2$). The inclusivity increases with an increase in the $\tau_2$, and it is higher for fewer keypoints.}
\label{fig:inclusivity_for_tau}
\end{figure}

The inclusivity metric (defined by~\cite{ref:clara_2020}) depends on the total number of keypoints and the tolerance threshold $\tau_2$. 
To validate this, we train our network separately for different numbers of keypoints, and calculate the inclusivity for different $\tau_2$. 
It is found that the inclusivity is higher for fewer keypoints, and it increases with the increase in the $\tau_2$. 
Fig.~\ref{fig:inclusivity_for_tau} shows the average inclusivity (of the test set) for different values of $\tau_2$. 

\subsection{Impact of the separation and shape loss on perturbation}
We observed that the separation and shape loss contribute to robustness, especially in the case of perturbation. To validate this, we have trained SC3K (for airplane category) without separation and shape loss and tested it for noisy and down-sampled PCDs. The results are illustrated in Tab.~\ref{tab:sup_kp_verrices}.
The performance drops significantly with an increase in the noise ratio or down-sampling scale.
\begin{table}[t]
\centering
\begin{adjustbox}{width=\columnwidth}
\begin{tabular}{c|c|ccc}
\toprule
\multicolumn{2}{c|}{\textbf{Noise/down-sample scale}}                       & \textbf{0/0x} & \textbf{0.02/4x} & \textbf{0.06/16x} \\ 
\midrule
\multirow{2}{*}{\textbf{All loss}}    & \multicolumn{1}{c|}{\textbf{Inc.}} & 87.2/87.2     & 83.1/84.9        & 79.6/82.6         \\
                                      & \multicolumn{1}{c|}{\textbf{Cov.}} & 96.3/96.3     & 95.2/96.1        & 92.6/95.6         \\ 
                                      \midrule
\multirow{2}{*}{\textbf{Without two losses}} & \multicolumn{1}{c|}{\textbf{Inc.}} & 80.2/80.2     & 74.9/60.1        & 59.2/50.8         \\
                                      & \multicolumn{1}{c|}{\textbf{Cov.}} & 90.9/90.9     & 87.8/86.0        & 84.5/82.2         \\ \bottomrule
\end{tabular}
\end{adjustbox}
\caption{Results for PCD perturbations without using separation and shape loss}
\label{tab:sup_kp_verrices}
\end{table}

\subsection{Impact of residual blocks}
We evaluated SC3K by replacing the residual blocks with Conv1D layers obtaining a performance decrease of
-3.54\% 	-8.21\% 	-6.70\% 
on inclusivity, convergence and DAS on keypointNet dataset. Such a remarkable drop supports our design choice.  

\subsection{Evaluation for different augmentation strategies} 
As reported in Tab.~\ref{tab:diff_augmentations}, we evaluate the effect of different augmentations during training, such as downsample (DS), Noise (N) and Rotation (R). 
We used the same test set (canonical, without noise and down-sampling) for evaluation.
Tab.~\ref{tab:diff_augmentations} shows that the coverage and inclusivity decrease due to augmentations, whereas DAS increases when we use noisy and downsampled train samples. We observed that for noisy downsampled samples, although the keypoints are semantically consistent, they are not estimated on the object's surface. 
Considering the average value, we suggest that only rotations should be used as augmentation for training.
\begin{table}[h]
\centering
\begin{adjustbox}{width=\columnwidth}
\begin{tabular}{p{2.5cm}|cccc}
\toprule
\textbf{Augmentation}  & \textbf{Inclusivity} & \textbf{Coverage} & \textbf{DAS} & \textbf{Average}  \\ \midrule
R           &  \textbf{75.89} & \textbf{95.63} & 69.71  & \textbf{80.41 }  \\
R + DS      &  73.61 & 	89.56 & 	71.72 & 	78.30        \\
R + N       & 70.78	& 91.11	& 65.98	& 75.95  \\
R + DS + N  & 70.47	 & 89.68  & \textbf{77.68 } & 	79.27  \\ 
\bottomrule
\end{tabular}
\end{adjustbox}
\caption{Evaluation of SC3K for different augmentations.}
\label{tab:diff_augmentations}
\end{table}

\section{Quantitative results}
This section presents some additional quantitative comparisons of SC3K with SOTA approaches.

\subsection{Comparison with USEEK~\cite{ref:xue2022useek}}
We compare our keypoints for random-oriented PCDs with those of USEEK. We select the four categories (airplane, chair, guitar and knife) for which the USEEK's pretrained weights are available. The results are reported in Tab.~\ref{tab:useek_vs_sc3k}. We found that, on average, SC3K's inclusivity and coverage are +16.71\% and +23.82\% higher than those of USEEK. 
However, USEEK estimates only 3 and 4 keypoints for knife and guitar, respectively, so those are well separated and semantically ordered. 
Moreover, considering pose estimation as a downstream task, we use the keypoints estimated by both methods to compute a relative pose between two randomly oriented PCDs. The mean/median of the pose error of SC3K is +19.75$^\circ$/+5.69$^\circ$ better than that of USEEK. Thanks to the mutual loss components for enabling SC3K to estimate aligned keypoints irrespective of orientation.
\begin{table}[h]
\centering
\begin{adjustbox}{width=\columnwidth}
\begin{tabular}{l|ccc|cc}
\toprule
\multicolumn{1}{c|}{\multirow{2}{*}{\textbf{Approach}}} & \multirow{2}{*}{\textbf{Inclusivity} $\uparrow$} & \multirow{2}{*}{\textbf{Coverage} $\uparrow$} & \multirow{2}{*}{\textbf{DAS} $\uparrow$} & \multicolumn{2}{c}{\textbf{Pose error} $\downarrow$} \\
\multicolumn{1}{c|}{}                                   &                                       &                                    &                               & \textbf{Mean}     & \textbf{Median}     \\
\midrule
USEEK                                                  & 80.49                                 & 73.42                              & 77.78                         & 30.12             & 6.45                \\
SC3K                                                   & 97.20                                 & 97.24                              & 72.70                         & 10.72             & 0.76                \vspace{0.15cm}\\
Difference                                             & +16.71                                 & +23.82                              & -5.08                         & +19.40             & +5.69        \\
\bottomrule
\end{tabular}
\end{adjustbox}
\caption{Comparison of SC3K with USEEK~\cite{ref:xue2022useek}. The keypoints estimated by SC3K are comparatively more useful for computing relative pose between two transformed versions of the same object.  
}
\label{tab:useek_vs_sc3k}
\end{table}

\subsection{Comparison with ISS~\cite{ref:zhong2009intrinsic} and MR~\cite{ref:yuan2022unsupervised}}
In this section, we compare DAS of SC3K with the other baseline approaches, i.e., ISS~\cite{ref:zhong2009intrinsic} and MR~\cite{ref:yuan2022unsupervised}. We consider their DAS exactly the same as reported in~\cite{ref:yuan2022unsupervised}. Our results are the same as we have reported in the main paper. However, we only show the DAS of the same categories as given in~\cite{ref:yuan2022unsupervised}. 
It can be observed that, on average, SC3K outperforms the other approaches.
\begin{table}[t]
\begin{center}
\begin{adjustbox}{width=\columnwidth}
\begin{tabular}{cccccc}
\toprule
         & ULCS~\cite{ref:clara_2020}  & SM~\cite{ref:skeleton_marger}  & ISS~\cite{ref:zhong2009intrinsic}   & MR~\cite{ref:yuan2022unsupervised}  & SC3K           \\
\midrule
Airplane &  61.40 & 77.70          & 13.10 & \underline{81.00}          & \textbf{82.86} \\
Chair    &  64.30 & 76.80          & 10.70 & \underline{83.10}         & \textbf{87.04} \\
Car      &  --     & \textbf{79.40} & 8.00  & 74.00          & \underline{75.19}         \\
Table    &  --     & 70.00          & 16.20 & \textbf{78.50} & \underline{76.03}         \\
Guitar   &  --     & \underline{63.10}          & 8.70  & 61.30          & \textbf{65.67} \\
Mug      &  --     & 67.20          & 11.20 & \underline{68.20}          & \textbf{79.25} \\
Cap      &  --     & 53.00          & 13.10 & \underline{57.10}          & \textbf{59.72} \vspace{0.1cm} \\ 
Mean     &  62.85 & 69.60          & 11.57 & \underline{71.89}          & \textbf{74.54} \\
\bottomrule
\end{tabular}
\end{adjustbox}
\end{center}
\caption{Comparison based on the semantic consistency between the keypoints estimated for different objects of the same category. The baseline results (DAS) are the same as reported in~\cite{ref:yuan2022unsupervised}. The higher value is best.
}
\label{tab:DAS_comparison}
\end{table}

\section{Qualitative results}
This section presents a qualitative comparison of the SC3K with the SOTA approaches. Moreover, it also shows the keypoints estimated by SC3K for intra-class, noisy and down-sampled objects.

\subsection{Comparison with ULCS~\cite{ref:clara_2020} and SM~\cite{ref:skeleton_marger}}
\begin{figure}[h]
\centering
\includegraphics[height=5.5cm]{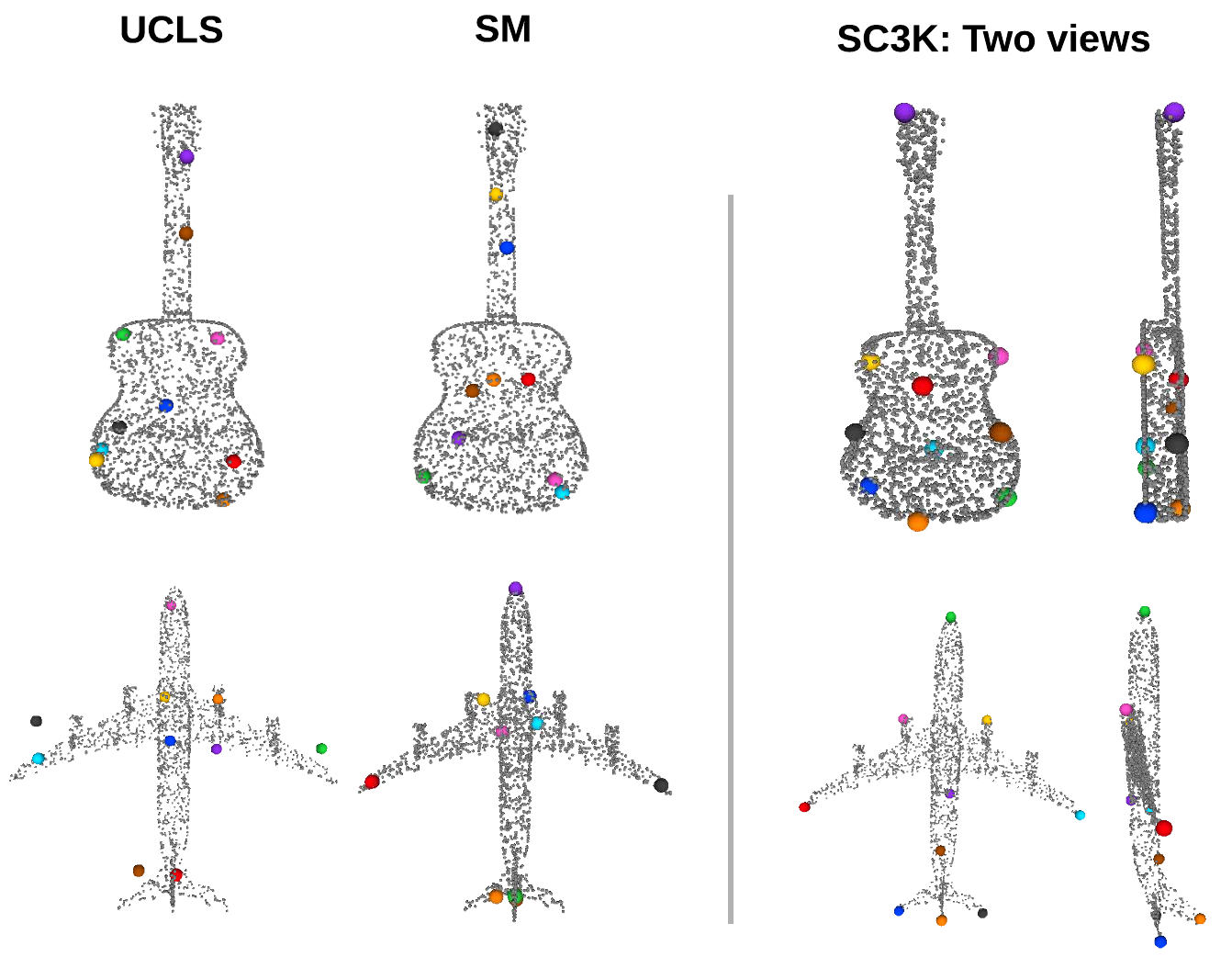}
\caption{Qualitative comparison. Columns 1 and 2 present keypoints estimated by ULCS and SM, respectively. Columns 3 and 4 show the keypoints estimated by SC3K.
It can be observed that some of keypoints of the ULCS are estimated outside the object (airplane). 
The keypoints estimated by SC3K best characterize the object's shape, as they are estimated on the surface and cover the complete object.}
\label{fig:qualitative_comparison}
\end{figure}

We show in this section the keypoints estimated by SC3K and  the SOTA approaches in Fig.~\ref{fig:qualitative_comparison}.
For better understanding, the estimated keypoints (in different colours) are shown on top of the original PCDs (in Gray).
The colour of the keypoints represents their semantic ID information, i.e. a point with the same colour should stay in the same area despite perturbations. Columns 1 and 2 illustrate the keypoints estimated by ULCS~\cite{ref:clara_2020} and SM~\cite{ref:skeleton_marger}, respectively. In contrast, two views of the keypoints estimated by the proposed approach are depicted in columns 3 and 4.
The comparison validates that our keypoints are estimated close to the surface, highlighting the corners, thus best characterizing the object's shape.

\subsection{Qualitative comparison with Intra-class objects}
\begin{figure}[h]
\centering
\includegraphics[height=18.0cm]{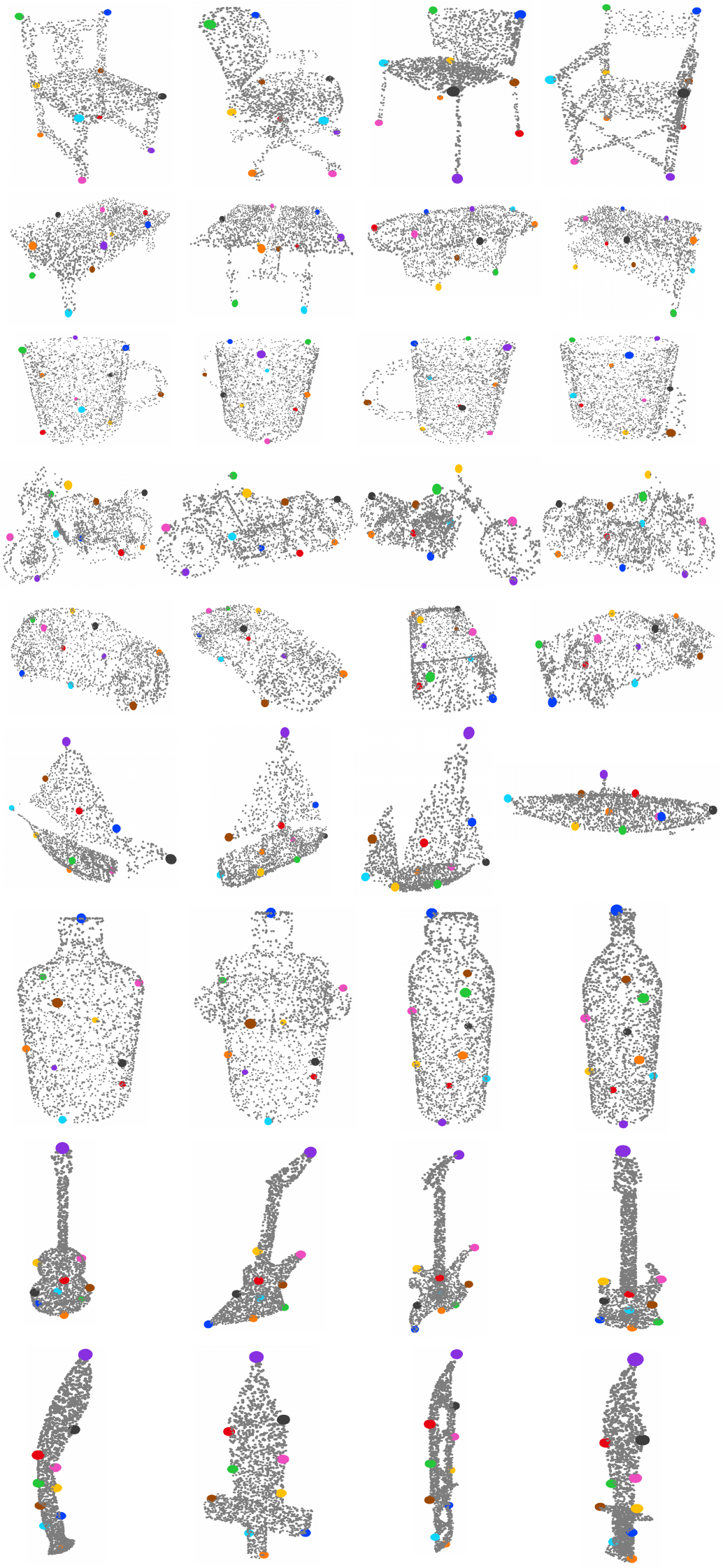}
\caption{Qualitative results of the proposed SE3K for different categories. Every row shows four objects (in different poses) of the same category. The keypoints (coloured points) are estimated on the surface and in the same pose as the pose of the original PCDs (small gray points). Moreover, they are semantically consistent for all the intra-class objects.}
\label{fig:intra-class_comparision}
\end{figure}

This section shows the qualitative results of the SC3K for intra-class objects. Four objects (in random poses) of the different categories are shown in Fig.~\ref{fig:intra-class_comparision}. 
It can be observed that the keypoints are proximal to the original PCDs, semantically in order (coherent) and pointing to the sharp edges of the objects. 

\subsection{Visualisation of the noisy PCDs}
\begin{figure*}[]
\centering
\includegraphics[height=20.0cm]{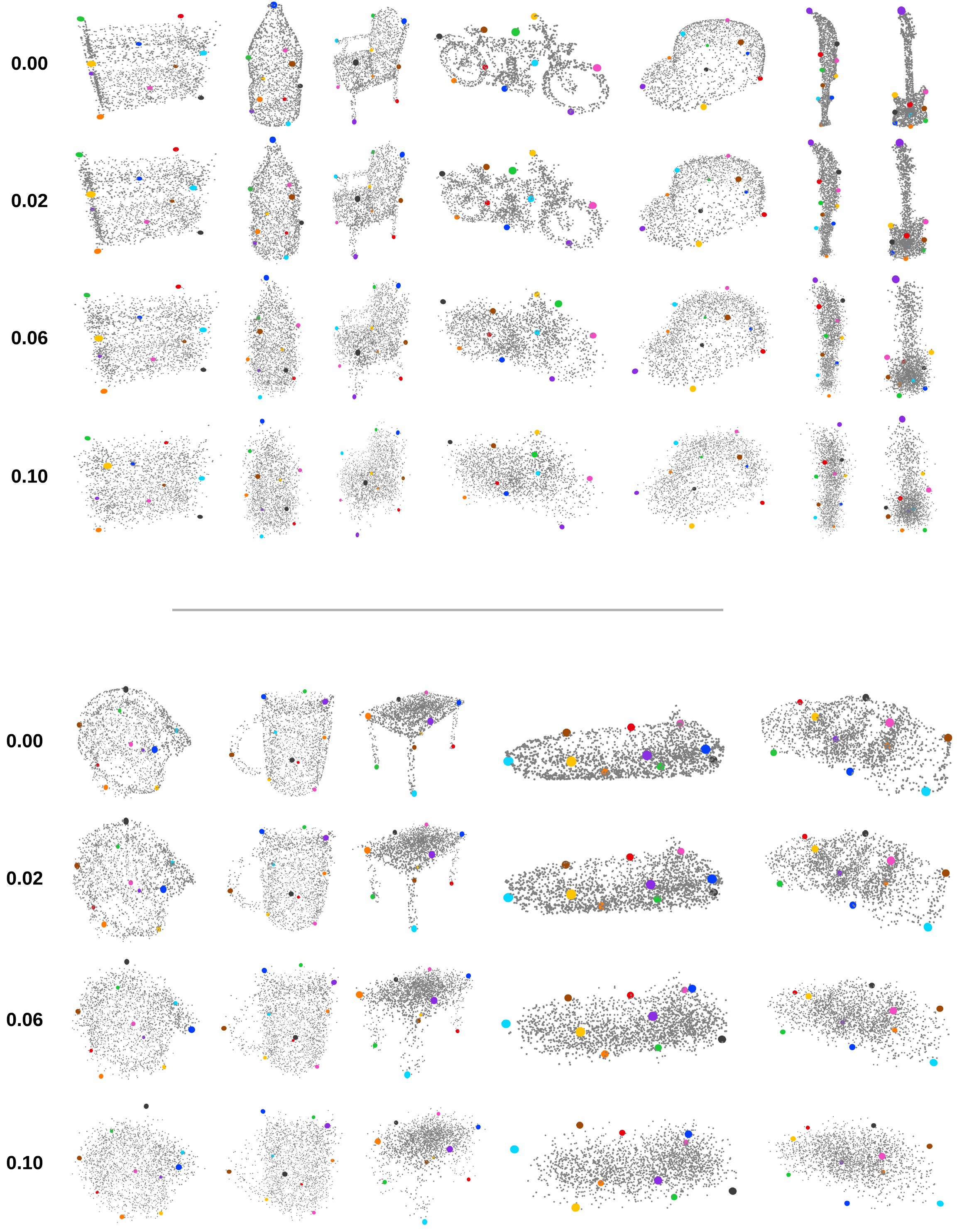}
\caption{Performance of the proposed approach for the noisy PCDs. Gaussian noise of different scales (as mentioned at the beginning of every row) is added to the input PCDs. ``0.00'' represents the original PCD (without noise). The SC3K remains successful in estimating the semantically consistent keypoints for noisy PCDs. However, the accuracy has decreased with an increase in the noise scale.}
\label{fig:noise_comparison_visualizations}
\end{figure*}

This section shows the qualitative results (extension of Fig.~\ref{fig:noisy_downsample_comparison}a in the main paper) of SC3K for different noisy PCDs. We add the Gaussian noise of different scales to the original PCDs of different categories. The noise scale is written in the beginning of every row where ``0.00'' mean original PCD without noise. 
The estimated keypoints are shown in Fig.~\ref{fig:noise_comparison_visualizations}. It can be observed that the proposed SC3K remains successful in estimating the 3D keypoints from the noisy PCDs. Moreover, the keypoints are always estimated close to the outermost points in the PCDs (i.e. close to the noisy surface). However, the accuracy decreases with the increase in the noise scale.

\subsection{Visualisation of the Down-sampled PCDs}

\begin{figure*}[]
\centering
\includegraphics[height=20.0cm]{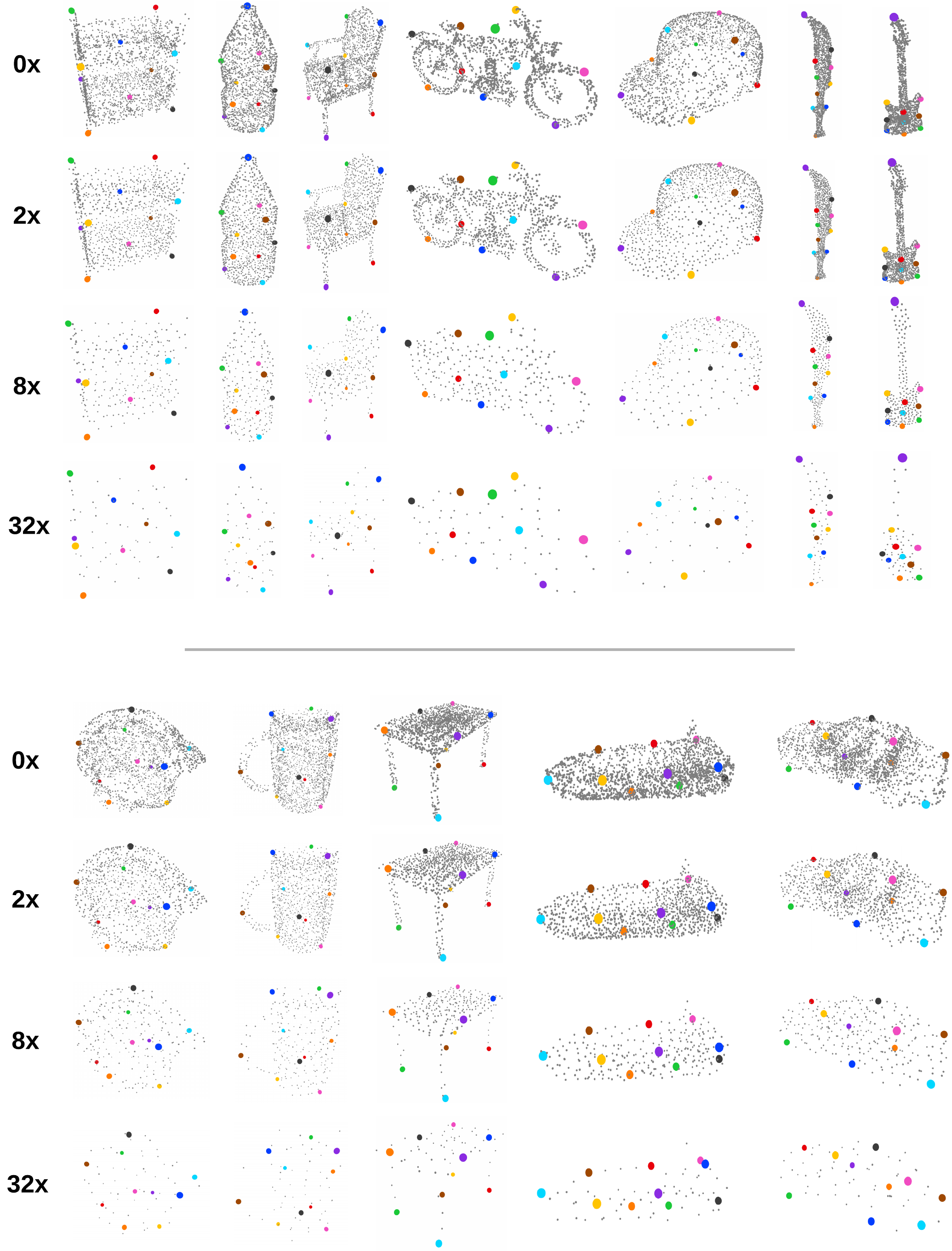}
\caption{Performance of our method for down-sampled PCDs. The input PCDs are down-sampled for different scales, as mentioned at the beginning of every row. The ``$0\times$'' shows the original PCDs. The proposed SC3K remains successful in estimating the approximately accurate 3D positions of the keypoints.}
\label{fig:downsample_comparison_visualizations}
\end{figure*}

This section presents the performance of SC3K for down-sampled PCDs as an extension of the results shown in Fig.~\ref{fig:noisy_downsample_comparison}b of the main paper. 
For decimating the PCD, we use the Farthest Point Sampling (FPS) as used in~\cite{ref:yuan2018pcn} to sample points from original PCDs for different sampling ratios. 
We test our pre-trained network to estimate the 3D keypoints from the down-sampled PCDs.
The results are shown in Fig.~\ref{fig:downsample_comparison_visualizations}.  
The figure is horizontally divided to fix all the objects on one page. 
Each column presents the results of a different object. 
The sampling ratio is shown at the beginning of every row. The ``$0\times$'' shows the original PCD without sampling (zero times sampling). 
It can be observed that the SC3K has estimated approximately accurate keypoints for the down-sampled PCDs. However, the keypoints are not estimated at the same positions as the positions of the corresponding keypoints of the original PCDs (without sampling) when the PCDs are scaled 32 times ($32\times$). The $32\times$ sampling means a PCD containing only 64 points, considering that the original PCD contains 2048 points.  

%
%
%
%
\end{document}